\documentclass[acmsmall,screen]{acmart}

\AtBeginDocument{%
  }

\setcopyright{acmcopyright}
\copyrightyear{2018}
\acmYear{2018}
\acmDOI{XXXXXXX.XXXXXXX}

\acmJournal{JACM}
\acmVolume{37}
\acmNumber{4}
\acmArticle{111}
\acmMonth{8}

\usepackage{multirow}
\usepackage{subcaption}
\usepackage{wrapfig}
\def\eg{\emph{e.g.}} 
\def\ie{\emph{i.e.}}

\def \mbf{\mathbf}
\def \mbb{\mathbb}
\def \tbf{\textbf}

\begin{document}

\title{Improving Continuous Sign Language Recognition with Consistency Constraints and Signer Removal}

\author{Ronglai Zuo}
\email{rzuo@cse.ust.hk}
\orcid{0000-0002-7184-5137}
\author{Brian Mak}
\email{mak@cse.ust.hk}
\orcid{0000-0001-6787-5555}
\affiliation{%
  \institution{The Hong Kong University of Science and Technology}
  \country{Hong Kong}
}








\renewcommand{\shortauthors}{Ronglai Zuo and Brian Mak}

\begin{abstract}
Deep-learning-based continuous sign language recognition (CSLR) models typically consist of a visual module, a sequential module, and an alignment module. However, the effectiveness of training such CSLR backbones is hindered by limited training samples, rendering the use of a single connectionist temporal classification loss insufficient. To address this limitation, we propose three auxiliary tasks to enhance CSLR backbones.
First, we enhance the visual module, which is particularly sensitive to the challenges posed by limited training samples, from the perspective of consistency. Specifically, since sign languages primarily rely on signers' facial expressions and hand movements to convey information, we develop a keypoint-guided spatial attention module that directs the visual module to focus on informative regions, thereby ensuring spatial attention consistency.
Furthermore, recognizing that the output features of both the visual and sequential modules represent the same sentence, we leverage this prior knowledge to better exploit the power of the backbone. We impose a sentence embedding consistency constraint between the visual and sequential modules, enhancing the representation power of both features.
The resulting CSLR model, referred to as consistency-enhanced CSLR, demonstrates superior performance on signer-dependent datasets, where all signers appear during both training and testing. To enhance its robustness for the signer-independent setting, we propose a signer removal module based on feature disentanglement, effectively eliminating signer-specific information from the backbone.
To validate the effectiveness of the proposed auxiliary tasks, we conduct extensive ablation studies. Notably, utilizing a transformer-based backbone, our model achieves state-of-the-art or competitive performance on five benchmarks, including PHOENIX-2014, PHOENIX-2014-T, PHOENIX-2014-SI, CSL, and CSL-Daily. Code and models are available at \href{https://github.com/2000ZRL/LCSA_C2SLR_SRM}{https://github.com/2000ZRL/LCSA\_C2SLR\_SRM}.
\end{abstract}


\received{20 February 2007}
\received[revised]{12 March 2009}
\received[accepted]{5 June 2009}

\maketitle

\section{Introduction}
\label{sec:intro}
Sign language is usually the principal communication method among hearing-impaired people.
Sign language recognition (SLR) aims to transcribe sign languages into glosses (basic lexical units in a sign language), which is an important technology to bridge the communication gap between the normal-hearing and hearing-impaired people.
According to the number of glosses in a sign sentence, SLR can be categorized into (a) isolated SLR (ISLR), in which each sign sentence consists of only a single gloss, and (b) continuous SLR (CSLR), in which each sign sentence may consist of multiple glosses. 
ISLR can be seen as a simple classification task, which becomes less popular in recent years.
In this paper, we focus on CSLR which is more practical than its isolated counterpart.
In recent years, more and more CSLR models are built using deep learning techniques because of their superior performance over traditional methods \cite{stmc, vac, sfl}.
According to \cite{sfl}, the backbone of most deep-learning-based CSLR models is composed of three parts: a visual module, a sequential (contextual) module, and an alignment module.
Within this framework, visual features are first extracted from sign videos by the visual module.
After that, sequential and contextual information are modeled by the sequential module.
Finally, due to the difference between the length of a sign video and its gloss label sequence, an alignment module is needed to align the sequential features with the gloss label sequence and yields its probability.

\begin{figure}[t]
  \centering
   \includegraphics[width=0.7\linewidth]{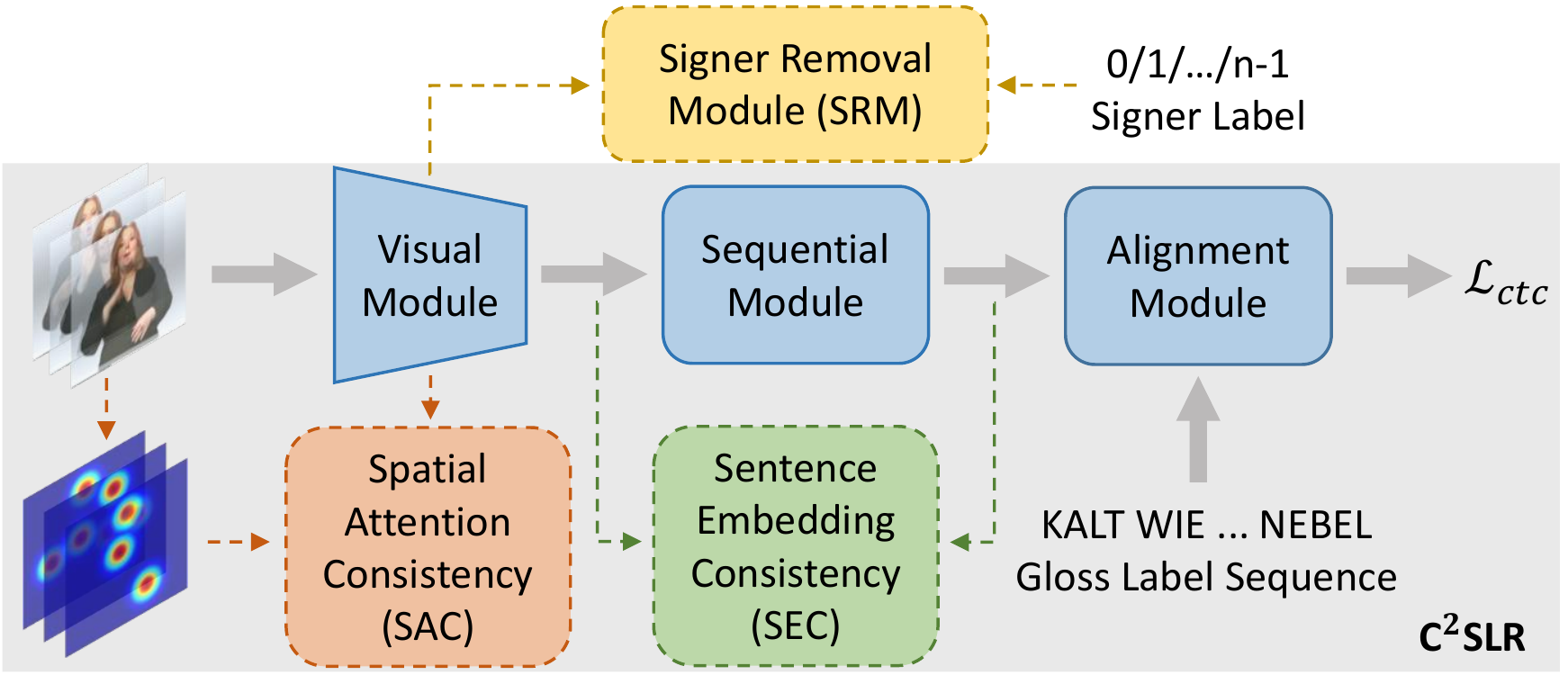}
   \caption{An overview of the CSLR backbone and the three proposed auxiliary tasks. First, our SAC enforces the visual module to focus on informative regions by leveraging pose keypoints heatmaps. Second, our SEC aligns the visual and sequential features at the sentence level, which can enhance the representation power of both the features simultaneously. SAC and SEC constitute our preliminary work \cite{zuo2022c2slr}, consistency-enhanced CSLR ($\text{C}^2$SLR). In this work, we extend $\text{C}^2$SLR by developing a novel signer removal module based on feature disentanglement for signer-independent CSLR.}
   \label{fig:intro}
\end{figure}

Usually, such CSLR backbones are trained with the connectionist temporal classification (CTC) \cite{ctc} loss.
However, since CSLR datasets are usually small, only using the CTC loss may not train the backbones sufficiently \cite{iopt, dnf, cma, stmc, self-mutual, fcn, vac}.
That is, the extracted features are not representative enough to be used to produce accurate recognition results.
To relieve this issue, existing works can be roughly divided into two categories.
First, \cite{dnf} proposes a stage optimization strategy to iteratively refine the extracted features with the help of pseudo labels, which is widely adopted in \cite{dilated, iopt, cma, stmc, self-mutual}.
However, it introduces more hyper-parameters and is time-consuming since the model needs to adapt to a different objective in each new stage \cite{fcn}.
As an alternative strategy, auxiliary learning can keep the whole model end-to-end trainable by just adding several auxiliary tasks \cite{fcn, vac}.
In this work, three novel auxiliary tasks are proposed to help train CSLR backbones.

Our first auxiliary task aims to enhance the visual module, which is important to feature extraction but sensitive to the insufficient training problem \cite{vac, dnf, stmc}.
Since the information of sign languages is mainly included in signers' facial expressions and hand movements \cite{stmc, koller2020quantitative, hu2021global}, signers' face and hands are treated as informative regions.
Thus, to enrich the visual features, some CSLR models \cite{stmc, papadimitriou20_interspeech} leverage an off-the-shelf pose detector \cite{cao2019openpose, sun2019deep} to locate the informative regions and then crop the feature maps to form a multi-stream architecture.
However, this architecture will introduce extensive parameters since each stream processes its inputs independently and the cropping operation may overlook the rich information in the pose keypoints heatmaps.
As shown in Figure \ref{fig:intro}, by visualizing the heatmaps, we find that they can reflect the importance of different spatial positions, which is similar to the idea of spatial attention.
Thus, as shown in Figure \ref{fig:framework}, we insert a lightweight spatial attention module into the visual module and enforce the spatial attention consistency (SAC) between the learned attention masks and pose keypoints heatmaps.
In this way, the visual module can pay more attention to the informative regions.

Only enhancing the visual module may not fully exploit the power of the backbone.
According to \cite{vac, self-mutual}, better performance can be obtained by explicitly enforcing the consistency between the visual and sequential modules.
VAC \cite{vac} adopts a knowledge distillation loss between the two modules by treating the visual and sequential modules as a student-teacher pair.
With a similar idea, SMKD \cite{self-mutual} transfers knowledge by shared classifiers.
Knowledge distillation can be treated as a kind of consistency since it is usually instantiated as the KL-divergence loss, a measurement of the distance between two probability distributions.
Nevertheless, the above two methods have a common deficiency that they measure consistency at the frame level, \ie, each frame has its own probability distribution.
We think that it is inappropriate to enforce frame-level consistency since the sequential module is supposed to gather contextual information; otherwise, the sequential module may be dropped.
Motivated by that both the visual and sequential features represent the same sentence, we propose the second auxiliary task: enforcing the sentence embedding consistency (SEC) between them.
As shown in Figure \ref{fig:framework}, we build a lightweight sentence embedding extractor that can be jointly trained with the backbone, and then minimize the distance between positive sentence embedding pairs while maximizing the distance between negative pairs.

We name the CSLR model trained with SAC and SEC as consistency-enhanced CSLR ($\text{C}^2$SLR).
According to our experimental results (Table \ref{tab:SD}), with a transformer-based backbone, $\text{C}^2$SLR can achieve satisfactory performance on signer-dependent datasets, in which all signers in the test set appear in the training set.
However, as shown in Table \ref{tab:2014SI}, $\text{C}^2$SLR cannot outperform the state-of-the-art (SOTA) work on the more challenging but realistic signer-independent CSLR (SI-CSLR).
Under the SI setting, since the signers in the test set are unseen during training, removing signer-specific information can make the model more robust to signer discrepancy.
In this work, we further develop a signer removal module (SRM) based on the idea of feature disentanglement.
More specifically, we first extract robust sentence-level signer embeddings with statistics pooling \cite{snyder2018x} to ``distill" signer information, which is then dispelled from the backbone implicitly by a gradient reversal layer \cite{ganin2016domain}.
Finally, the SRM is trained with a signer classification loss.
To the best of our knowledge, we are the first to develop a specific module for SI-CSLR\footnote{Some works \cite{dnf, cma} evaluate their methods on SI-CSLR datasets, but none of them propose any dedicated modules for the SI setting. \cite{yin2016iterative} proposes a metric learning method to deal with the SI situation, but it focuses on ISLR.}.

In summary, our main contributions are:
\begin{itemize}
    \item We propose to enforce the consistency between the learned attention masks and pose keypoints heatmaps to enable the visual module to focus on informative regions.
    \item We propose to align the visual and sequential features at the sentence level to enhance the representation power of both features simultaneously.
    \item We propose a signer removal module from the idea of feature disentanglement to implicitly remove signer information from the backbone for SI-CSLR. To the best of our knowledge, we are the first to focus on this challenging setting.
    \item Extensive experiments are conducted to validate the effectiveness of the three auxiliary tasks. More remarkably, with a transformer-based backbone, our model can achieve SOTA or competitive performance on five benchmarks, while the whole model is trained in an end-to-end manner.
\end{itemize}

This work is an extension to our 2022 CVPR paper, $\text{C}^2$SLR \cite{zuo2022c2slr}. More specifically, we make the following new contributions:
\begin{itemize}
    \item Besides the investigation on signer-dependent continuous sign language recognition (SD-CSLR) in the CVPR paper, we propose in this paper an additional signer removal module (SRM) to tackle the more challenging signer-independent continuous sign language recognition (SI-CSLR) problem. More specifically, the SRM is designed to remove signer information from the backbone for SI-CSLR based on feature disentanglement. To the best of our knowledge, we are the first to propose a dedicated module to deal with SI-CSLR.
    \item We successfully adapt statistics pooling to SI-CSLR to extract robust sentence-level signer embeddings for the SRM.
    \item We conduct sufficient ablation studies to validate the effectiveness of the SRM, and the combination of $\text{C}^2$SLR and SRM can achieve SOTA performance on an SI-CSLR benchmark.
    \item We also report additional experimental results of $\text{C}^2$SLR on the latest large-scale Chinese sign language dataset, CSL-Daily \cite{zhou2021improving} with a vocabulary size of 2K and about 20K videos.
\end{itemize}

\section{Related Works}
\subsection{Deep-learning-based CSLR}
According to \cite{sfl}, most deep-learning-based CSLR backbones consist of a visual module (3D-CNNs \cite{iopt, csl-3} or 2D-CNNs \cite{stmc, vac, self-mutual}), a sequential module (1D-CNNs \cite{dense, fcn}, RNNs \cite{stmc, vac, self-mutual, iopt, cma}, or Transformer \cite{sfl, slt}), and an alignment module (CTC \cite{stmc, vac, self-mutual} or hidden Markov models \cite{cnn-lstm-hmm}).
To mitigate the issue of insufficient training, a novel approach is introduced by \cite{dnf}, which employs a stage optimization strategy to iteratively refine the extracted features utilizing pseudo labels. This technique has garnered significant attention and has been widely embraced in relevant studies \cite{iopt, cma, stmc, self-mutual}. Extending the capabilities of this strategy, \cite{iopt} incorporates a Long Short-Term Memory (LSTM) based auxiliary decoder. Furthermore, SMKD \cite{self-mutual} proposes a comprehensive three-stage optimization strategy that necessitates training the model over 100 epochs.
VAC \cite{vac} proposes an approach that achieves improved training time efficiency while enhancing the visual module and enforcing consistency between the visual and sequential modules. This is accomplished through the proposed visual enhancement and visual alignment constraints on the frame-level probability distributions. Notably, the proposed model is end-to-end trainable.
In this work, we enhance the visual module from a novel view of spatial attention consistency, and align the two modules at the sentence level to enforce their sentence embedding consistency.

Recently published CSLR works mostly focus on injecting more domain knowledge into sign video modeling \cite{chentwo,hu2023continuous,Jiao_2023_ICCV}, better training techniques \cite{Guo_2023_CVPR, zheng2023cvt}, or cross-lingual signs \cite{Wei_2023_ICCV}.
However, all these works still focus on the signer-dependent setting, which limits their application scenario.
In this work, we propose a signer removal module to make the model robust to signer discrepancy in the more realistic signer-independent setting.

\subsection{Spatial Attention}
Spatial attention mechanism allows models to selectively attend to specific positions, which is widely adopted in various computer vision tasks including semantic segmentation \cite{fu2019dual}, object detection \cite{woo2018cbam, cao2019gcnet}, and image classification \cite{woo2018cbam, cao2019gcnet, linsley2018learning}.
However, the spatial attention module may not be well-trained with a single task-specific loss function.
Leveraging external information to guide the spatial attention module can be a solution to this issue.
In \cite{chen2019motion}, the spatial attention module is guided by motion information for video captioning.
\cite{pang2019mask} and \cite{li2020relation} propose mask and relation guidance for occluded pedestrian detection and person re-identification, respectively.
GALA \cite{linsley2018learning} presents an intriguing approach by utilizing click maps obtained from a game as supervision. In this work, we leverage pose keypoints heatmaps to direct the learning process of the spatial attention module.

\subsection{Sentence Embedding}
Traditional methods \cite{palangi2016deep, liu2019cross} commonly adopt a straightforward approach where the word embedding sequence is directly fed into recurrent neural networks (RNNs), and the final hidden state (or two hidden states for bidirectional RNNs) is extracted as the sentence embedding.
Recently, many powerful sentence embedding extractors \cite{reimers2019sentence, gao2021simcse, carlsson2020semantic} are built on BERT \cite{kenton2019bert}.
However, it is difficult to use these methods in our work because (1) they are too large to be co-trained along with the backbone; (2) they are pretrained on spoken languages, which are totally different to sign languages represented by videos.
In this work, we build a lightweight sentence embedding extractor that can be jointly trained with the CSLR backbone.

\subsection{Feature Disentanglement}
\label{sec:disent}
In the context of signer-independent continuous sign language recognition (SI-CSLR), each signer can be considered as a distinct domain, and the key is to enable the model to generalize well to unseen
domains, \ie, the test signers.
Feature disentanglement has emerged as a powerful approach for achieving domain generalization by decomposing features into domain-invariant and domain-specific components \cite{wang2021generalizing}.
Adversarial learning has gained significant traction in the field of feature disentanglement, with the feature extractor serving as the generator and the domain classifier as the discriminator \cite{xu2020investigating, cheng2021puregaze, liu2018exploring}. For instance, in the context of facial expression recognition, \cite{xu2020investigating} employs an adversarial approach to mitigate biases such as gender and race by training a series of domain classifiers. \cite{cheng2021puregaze} introduces a self-adversarial framework specifically designed to remove gaze-irrelevant factors, resulting in improved gaze estimation performance. Another notable advancement in feature disentanglement is the utilization of attention mechanisms to emphasize task-relevant features, while considering the remaining features as task-irrelevant. This approach has been successfully employed in various domains. For instance, in person re-identification, \cite{jin2020style} utilizes a channel attention module to suppress style information, while in face recognition, \cite{huang2021age} incorporates both spatial and channel attention mechanisms to eliminate age-related features. These studies exemplify the efficacy of leveraging adversarial learning and attention mechanisms for feature disentanglement in a range of applications.
However, adversarial learning is usually complicated as the generator and discriminator are trained iteratively, and the attention modules would introduce extra parameters.
In this work, we adopt the gradient reversal (GR) layer \cite{ganin2016domain} that reverses the gradient coming from the domain (signer) classification loss when the back-propagation process arrives at the feature extractor (CSLR backbone) while keeping the gradient of the domain classifier unchanged.
It shares a similar idea with adversarial learning, but it is totally end-to-end and introduces no extra parameters compared to attention-based methods.
Thus, we believe it can serve as a simple baseline for future research on SI-CSLR.

\begin{figure*}[t]
  \centering
  \includegraphics[width=1.0\linewidth]{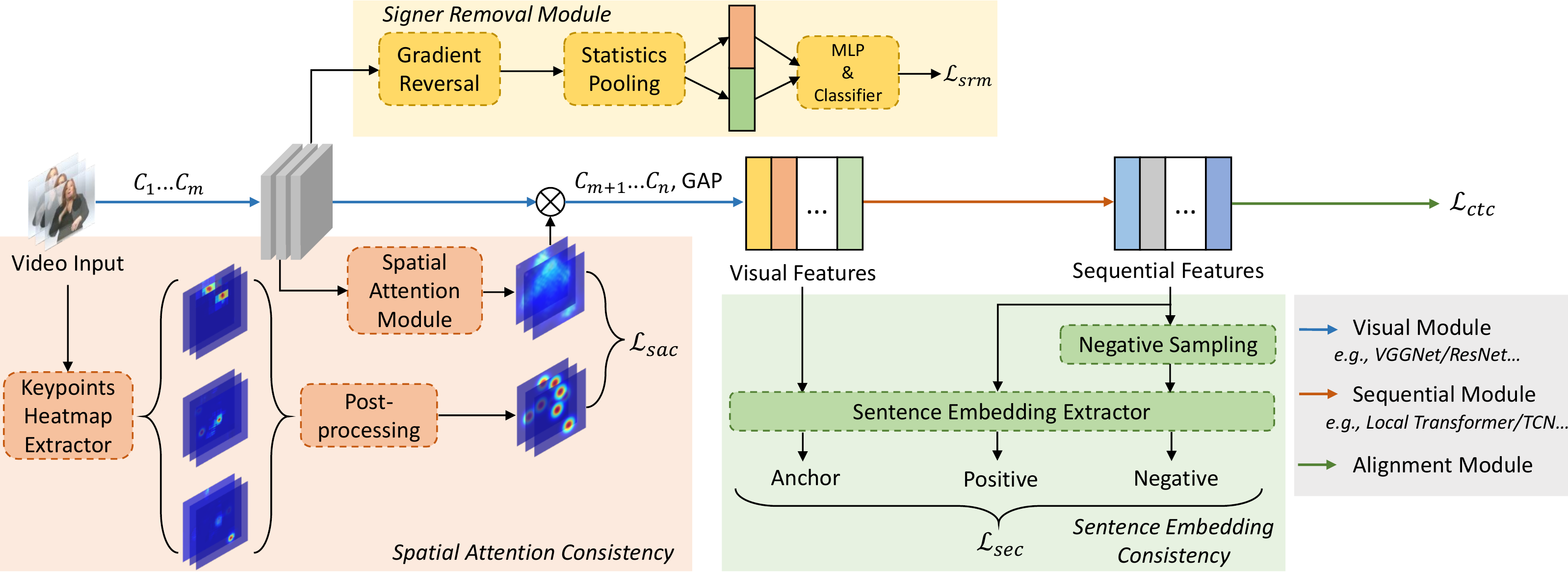}
  \caption{An overview of our proposed method. The sign video input is first fed into the visual module (\eg, VGGNet \cite{vggnet} and ResNet \cite{resnet}) to extract visual features. The following sequential module (\eg, local Transformer (see details in Section \ref{sec:lt}) and TCN) further models long-/short-term dependencies and yield sequential features. The CTC loss \cite{ctc} is adopted as the main objective function. Three auxiliary tasks (highlighted in different colors) are proposed to improve the performance of the CSLR backbone. For spatial attention consistency, we insert a keypoint-guided spatial attention module after the $m$-th convolution layer, $C_m$, of the visual module. Besides, we push the model to align visual and sequential features at the sentence level to enhance their representative power. Finally, we introduce a signer removal module to make the model more robust to signer discrepancy under the signer-independent setting.}
  \label{fig:framework}
\end{figure*}

\section{Our Proposed Method}
\subsection{Framework Overview}
\label{sec:overview}
Figure \ref{fig:framework} gives an overview of our proposed method. The blue, orange, and green arrows represent the three components of the CSLR backbone: visual module, sequential module, and alignment module, respectively.
Taking a sign video with $T$ RGB frames $\mbf{x}=\{\mbf{x}_t\}_{t=1}^T \in \mathbb{R}^{T\times H \times W \times 3}$ as input, the visual module, which simply consists of several 2D-CNN\footnote{We only consider visual modules based on 2D-CNNs since a recent survey \cite{survey} shows that 3D-CNNs cannot provide as precise gloss boundaries as 2D-CNNs, and lead to worse performance.} layers ($C_1, \dots, C_n$) followed by a global average pooling (GAP) layer, first extracts visual features $\mbf{v}=\{\mbf{v}_t\}_{t=1}^T \in \mathbb{R}^{T\times d}$.
The sequential features $\mbf{s}=\{\mbf{s}_t\}_{t=1}^T \in \mathbb{R}^{T\times d}$ will be further extracted by the sequential module.
Finally, the alignment module computes the probability of the gloss label sequence $p(\mbf{y}|\mbf{x})$ based on the widely-adopted CTC \cite{ctc}, where $\mbf{y}=\{y_i\}_{i=1}^N$ and $N$ denotes the length of the gloss sequence.
Below we will first present our proposed three auxiliary tasks, spatial attention consistency (Section \ref{sec:sac}), sentence embedding consistency (Section \ref{sec:sec}), and signer removal (Section \ref{sec:srm}), respectively. The overall loss function will be formulated in Section \ref{sec:loss}. Finally in Section \ref{sec:lt}, we will introduce a variant of Transformer as a strong sequential module for CSLR.

\subsection{Spatial Attention Consistency (SAC)}
\label{sec:sac}
Signers' facial expressions and hand movements are two major clues of sign languages \cite{koller2020quantitative, stmc, zuo2023natural}.
Thus, it is reasonable to expect the visual module can focus on signers' face and hands, \ie, informative regions (IRs).
From this view, we insert a spatial attention module into the visual module and enforce the consistency between the learned attention masks and keypoints heatmaps.
Since SAC is applied to all frames in the same way, we will omit the time steps in the formulation below.

\begin{figure}[t]
\begin{subfigure}[t]{.5\textwidth}
 \centering
 \includegraphics[width=\textwidth]{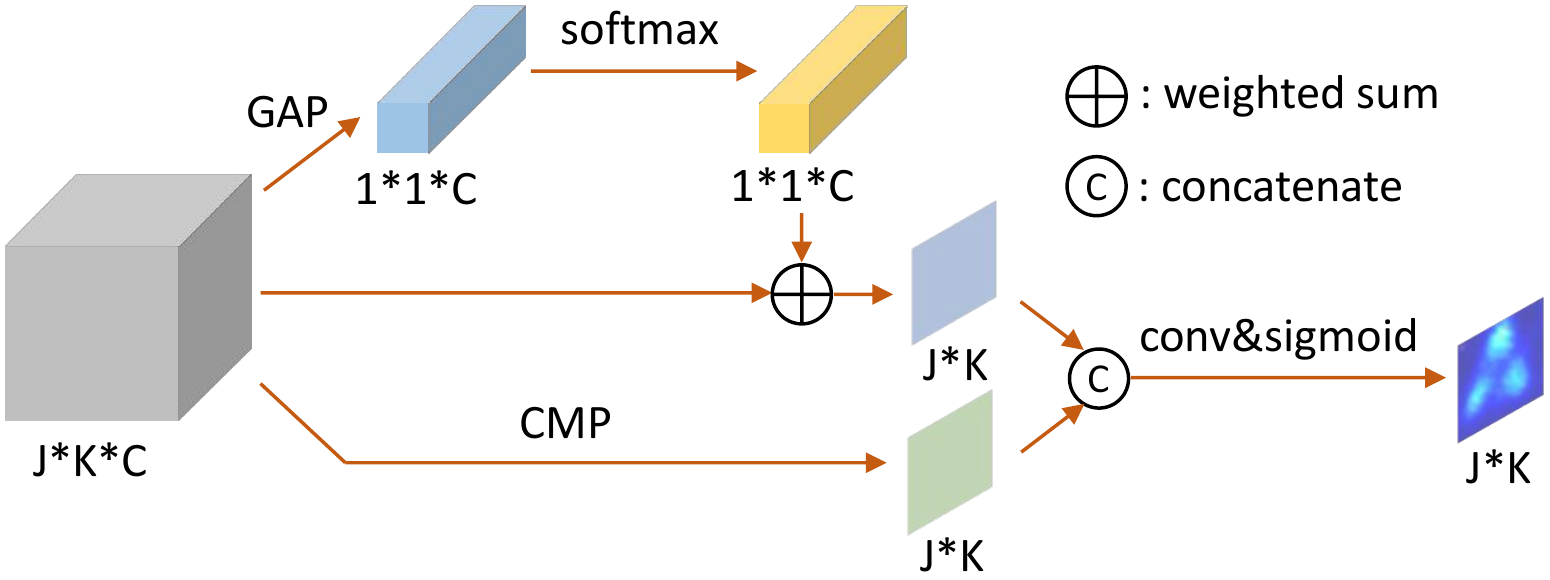}
 \caption{}
 \label{fig:sac}
\end{subfigure}
\hfill
\begin{subfigure}[t]{.45\textwidth}
  \centering
  \includegraphics[width=\textwidth]{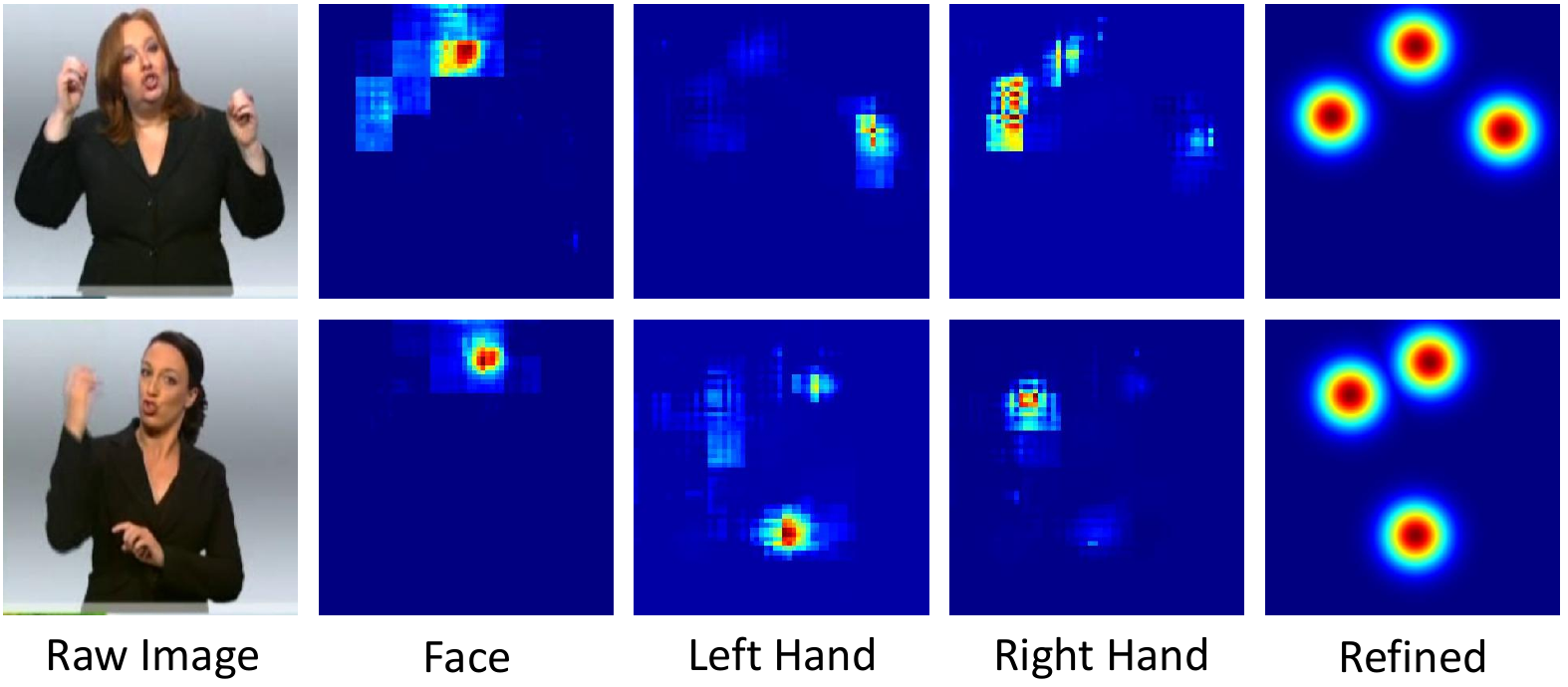}
  \caption{}
  \label{fig:heatmap}
\end{subfigure}
\caption{(a) The architecture of our spatial attention module. ($J\times K\times C$: the size of the input feature maps, GAP: global average pooling, CMP: channel-wise max pooling. (b) Two examples of the original and refined heatmaps.}
\end{figure}

\subsubsection{Spatial Attention Module}
We build our spatial attention module based on CBAM \cite{woo2018cbam} due to its simplicity and effectiveness.
As shown in Figure \ref{fig:sac}, we first pick the most informative channel via a channel-wise max pooling (CMP) operation:
\begin{equation}
    \mbf{M}_1 = f_{CMP}(\mbf{F}) \in \mathbb{R}^{J\times K\times 1},
\label{eq:1}
\end{equation}
where $\mbf{M}_1$ is the squeezed feature map by CMP, and $\mbf{F}\in\mathbb{R}^{J\times K \times C}$ is the input feature maps.

Besides CMP, CBAM also squeezes the feature maps with an average pooling operation along the channel dimension. 
However, we propose to dynamically weight the importance of each channel.
As shown in Figure \ref{fig:sac}, we first conduct global average pooling (GAP) over $\mbf{F}$ to gather global spatial information. 
Then the channel weights $\mbf{E}\in (0,1)^{1\times 1 \times C}$ are simply generated by a channel-wise softmax layer.
By a weighted sum along the channel dimension, we can generate another squeezed feature map $\mbf{M}_2$:
\begin{equation}
    \mbf{M}_2 = \mbf{F} \oplus \mbf{E} = \sum_{i=1}^C \mbf{F}_i \cdot \mbf{E}_i \in \mathbb{R}^{J\times K\times 1},
\end{equation}
Finally, the spatial attention mask $\mbf{M}$ is generated as:
\begin{equation}
    \mbf{M} = \sigma (f_{conv}(cat(\mbf{M}_1, \mbf{M}_2))) \in (0,1)^{J\times K},
\end{equation}
where $\sigma(\cdot)$ is the sigmoid function, $f_{conv}(\cdot)$ is a 2D-CNN layer with a kernel size of 7\texttimes7, and $cat(\cdot,\cdot)$ is a channel-wise concatenation operation. 
The output feature maps will be a product between $\mbf{F}$ and $\mbf{M}$.
In this way, important positions can be highlighted while trivial ones can be suppressed.

It should be noted that our channel weights are similar to the channel attention module in CBAM, but ours introduces no extra parameters and can even outperform the vanilla CBAM according to our ablation studies in Table \ref{tab:sac}.

\subsubsection{Keypoints Heatmap Extractor}
Simply training the spatial attention module with the backbone may lead to sub-optimal solutions.
Given the prior knowledge that signers' faces and hands are informative regions (IRs), we guide the spatial attention module with keypoints heatmaps extracted by the pretrained HRNet \cite{sun2019deep, andriluka20142d}.
Specifically, we first normalize the raw outputs of HRNet linearly to obtain the original heatmaps:
\begin{equation}
    \mbf{H}_o^i = \frac{f_H^i(\mbf{I}) - \min {f_H^i(\mbf{I})}} {\max {f_H^i(\mbf{I})} - \min{f_H^i(\mbf{I})}} \in [0,1]^{H\times W},
\end{equation}
where $\mbf{I}$ is the raw RGB frame, $f_H(\cdot)$ is the pretrained HRNet, and $i\in\{1,2,3\}$ denotes the face, left hand, and right hand, respectively.

\subsubsection{Post-processing}
\label{sec:post-process}
There are some defects in the original heatmaps although they can roughly highlight the positions of IRs.
As shown in Figure \ref{fig:heatmap}, some trivial regions, \eg, the top of the face heatmap in the first row and the middle part of the left hand heatmap in the second row, may receive high activation values.
Besides, some highlighted regions, \eg, both of the face heatmaps in Figure \ref{fig:heatmap}, may not cover the IRs entirely.
In addition, there is usually a mismatch between the fixed heatmap resolution of the pretrained HRNet and that of the spatial attention masks.
Below we will elaborate our heatmap post-processing module that corrects the mismatch.

We first locate the center of each IR from the original heatmaps via a simple argmax operation: $(x_i, y_i) =\mathrm{argmax}\ \mbf{H}_o^i$.
To fit different resolutions of spatial attention masks, we normalize the center as $(\hat{x}_i, \hat{y}_i) = (\frac{x_i}{H-1}, \frac{y_i}{W-1})$.
Suppose the spatial attention masks have a common resolution of $J \times K$, then a Gaussian-like refined keypoints heatmap is generated for each IR to reduce unwanted noise:
\begin{equation}
\label{equ:post}
    \mbf{H}_r^i(a,b) = \exp{\left(-\frac{1}{2}\left(\frac{(a-\hat{c}_i^x)^2}{(J/\gamma_x)^2}+\frac{(b-\hat{c}_i^y)^2}{(K/\gamma_y)^2}\right)\right)},
\end{equation}
where $0\leq a < J$, $0\leq b < K$. 
$(\hat{c}_i^x, \hat{c}_i^y) = (\hat{x}_i(J-1), \hat{y}_i(K-1))$, which denotes the transformed center for each IR under the resolution $J \times K$.
$\gamma_x$ and $\gamma_y$ are two hyper-parameters to control the scale of the highlighted regions.
In real practice, we set $\gamma_x=\gamma_y$.
Finally, we merge the three processed IR heatmaps into a single one: $\mbf{H}_r = \underset{i}{\max}\ \mbf{H}_r^i \in (0,1)^{J\times K}$.

\subsubsection{SAC Loss}
The spatial attention module is guided by the refined keypoints heatmaps via the SAC loss\footnote{For implementation, we further compute the average of $\mathcal{L}_{sac}$ over all time steps.}:
\begin{equation}
    \mathcal{L}_{sac} = \frac{1}{J\times K} \| \mbf{M}-\mbf{H}_r \|_2^2.
\end{equation}

\subsection{Sentence Embedding Consistency (SEC)}
\label{sec:sec}
\begin{wrapfigure}{r}{0.5\textwidth}
  \begin{center}
    \includegraphics[width=0.48\textwidth]{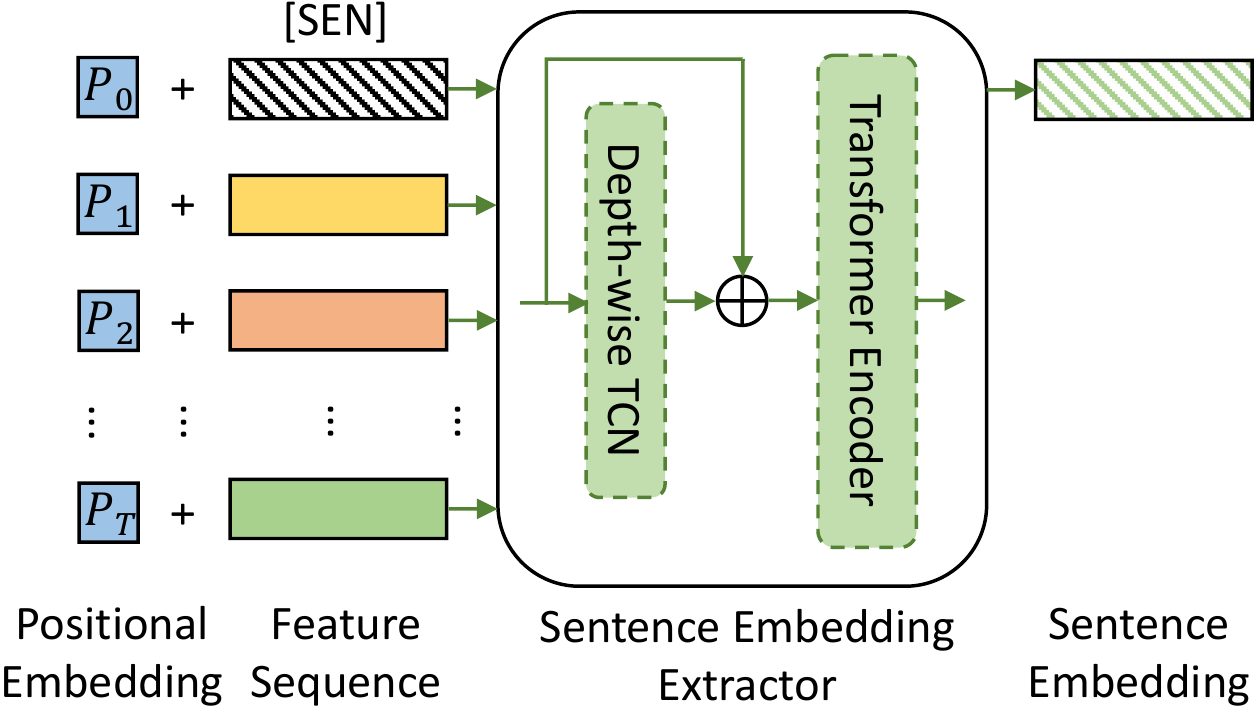}
  \end{center}
  \caption{The workflow of sentence embedding extraction. We omit LayerNorm \cite{layernorm} for simplicity.}
  \label{fig:sec}
\end{wrapfigure}

Some works \cite{vac, self-mutual} find that enforcing the consistency between the visual and sequential features can enhance their representation power, and lead to better performance.
Different from \cite{vac, self-mutual} that measure their consistency at the frame level, we impose a sentence embedding consistency between them.

\subsubsection{Sentence Embedding Extractor (SEE)}
Within a sign video, each gloss consists of only a few frames. 
We believe a good SEE for sign languages should take local contexts into consideration.
As shown in Figure \ref{fig:sec}, our SEE is built on QANet \cite{qanet}, which consists of a depth-wise temporal convolution network (TCN) layer and a transformer encoder layer.
The depth-wise TCN first extracts local contextual information from the frame-level feature sequence, then the transformer encoder models global contexts by its inner self-attention module.

Similar to the class token in BERT \cite{kenton2019bert}, we first prepend a learnable sentence embedding token, [SEN], to the sequential features $\mbf{s}\in\mathbb{R}^{T\times d}$ defined in Section \ref{sec:overview}:
\begin{equation}
    \mbf{s}' = cat(\text{[SEN]}, \mbf{s}) \in \mathbb{R}^{(T+1)\times d}.
\end{equation}
The input of the SEE is the summation of the feature sequence and the positional embeddings \cite{transformer}; \ie, $\mbf{s}''=\mbf{s}' + \mbf{P}$, where $\mbf{P}\in\mathbb{R}^{(T+1)\times d}$.

Within the SEE, the depth-wise TCN \cite{wu2018pay} layer first models local contexts with a residual shortcut: $\mbf{s}_l''=f_{TCN}(\mbf{s}'')+\mbf{s}''$. 
Then the transformer encoder layer gathers information from all time steps to get the sentence embedding:
\begin{equation}
    \mbf{E}_{sen}^s = f_{TF}(\mbf{s}_l'') \in \mathbb{R}^d.
\end{equation}
We can also get the sentence embedding of visual features, $\mbf{E}_{sen}^v$, in the same way.

\subsubsection{Negative Sampling}
Directly minimizing the distance between $\mbf{E}_{sen}^s$ and $\mbf{E}_{sen}^v$ will result in trivial solutions.
For example, if the parameters of SEE are all zeros, then the outputs of SEE will always be the same.
A simple way to address this issue is introducing negative samples.
In this work, we follow the common practice \cite{schroff2015facenet, ye2019unsupervised, oord2018representation, hjelm2018learning} and sample another video from the mini-batch and take its sequential features as the negative sample.
Note that most CSLR models \cite{vac, self-mutual, stmc} are trained with a batch size of 2, and our negative sampling strategy will degenerate to swapping under this setting:
\begin{equation}
    (neg(\mbf{B}[0]), neg(\mbf{B}[1])) = (\mbf{B}[1], \mbf{B}[0]),
\end{equation}
where $\mbf{B} \in \mathbb{R}^{2\times T\times d}$ is a mini-batch of the sequential features, and $neg(\mbf{B}[\cdot])$ denotes the corresponding negative sample.

\subsubsection{SEC Loss}
We implement SEC loss as a triplet loss \cite{schroff2015facenet} and minimize the distances between the sentence embeddings computed from the visual and sequential features of the same sentence, while maximizeing the distances between those from different sentences:
\begin{equation}
\begin{split}
    \mathcal{L}_{sec} = \max &\{d(\mbf{E}_{sen}^v, \mbf{E}_{sen}^s) - d(\mbf{E}_{sen}^v, neg(\mbf{E}_{sen}^{s}))+\alpha, 0\},
\end{split}
\label{equ:sec}
\end{equation}
where $d(\mbf{x}_1,\mbf{x}_2)=1-\frac{\mbf{x}_1 \cdot \mbf{x}_2}{\|\mbf{x}_1\|_2 \cdot \|\mbf{x}_2\|_2}$; 
$\{\mbf{E}_{sen}^v, \mbf{E}_{sen}^s\}$ are sentence embeddings of visual and sequential features from the same sentence;
$\{\mbf{E}_{sen}^v, neg(\mbf{E}_{sen}^{s})\}$ are sentence embeddings of visual and sequential features from different sentences, and we treat the sentence embedding of the sequential features from a different sentence as the negative sample $neg(\mbf{E}_{sen}^{s})$; $\alpha$ is the margin.

\subsection{Signer Removal Module (SRM)}
\label{sec:srm}
\begin{figure}[t]
  \centering
  \includegraphics[width=0.7\linewidth]{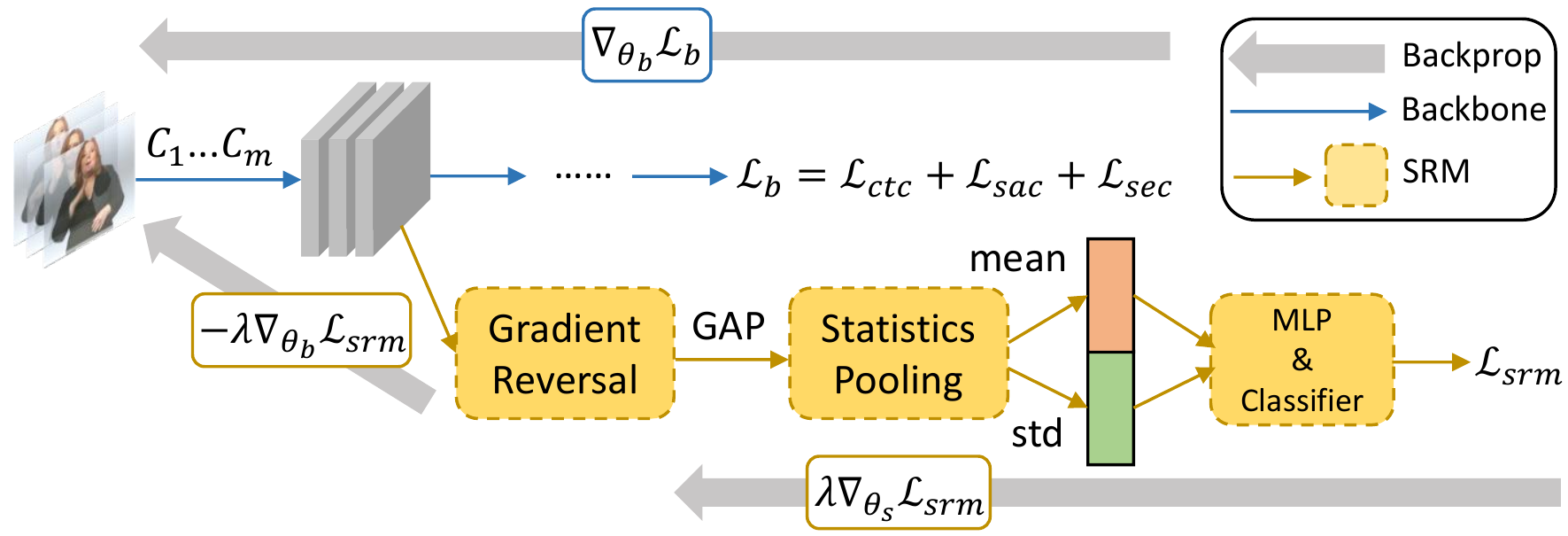}
  \caption{Workflow of our signer removal module (SRM). We insert the SRM after the $m$-th CNN layer, $C_m$. The loss of $\text{C}^2$SLR, $\mathcal{L}_b$, which is a sum of the CTC, SAC, and SEC losses, is used to train the backbone parameters $\theta_b$. The signer classification loss $\mathcal{L}_{srm}$ is used to train the SRM parameters $\theta_s$ as usual, while the gradient from $\mathcal{L}_{srm}$ will be reversed for $\theta_b$. $\lambda$ is the loss weight for $\mathcal{L}_{srm}$.}
  \label{fig:srm}
\end{figure}

To remove signer information from CSLR backbones, we further develop a signer removal module (SRM) based on statistics pooling and gradient reversal as shown in Figure \ref{fig:srm}. 

\subsubsection{Signer Embeddings}
We first extract signer embeddings to ``distill" signer information before dispelling it.
A na\"ive method is simply feeding the frame-level features into an MLP, and treat the outputs of MLP as signer embeddings.
In this work, motivated by the superior performance of x-vectors \cite{snyder2018x} in speaker recognition, we leverage statistics pooling to obtain more robust sentence-level signer embeddings.

Specifically, we first feed the intermediate visual features $\mbf{F} \in \mbb{R}^{T\times J \times K \times C}$ into a global average pooling layer to squeeze the spatial dimension and obtain frame-level features\footnote{Here we misuse the notation $\mbf{F}$ in Equation \ref{eq:1}.} $\mbf{F}_s \in \mbb{R}^{T\times C}$. 
Then a statistics pooling (SP) layer is used to aggregate frame-level information:
\begin{equation}
    \mbf{F}_s^{SP} = cat(\mbf{F}_s^{mean}, \mbf{F}_s^{std}) \in \mbb{R}^{2C},
\end{equation}
where $\mbf{F}_s^{mean} \in \mbb{R}^C$ and $\mbf{F}_s^{std} \in \mbb{R}^C$ are the temporal mean and standard deviation of $\mbf{F}_s$, respectively. 
In this way, $\mbf{F}_s^{SP}$ is capable to capture signer characteristics over the entire video instead of at the frame-level.

After that, a simple two-layer MLP with rectified linear unit (ReLU) function is used to project the statistics into the signer embedding space:
\begin{equation}
    \mbf{E}_{sig} = ReLU(\mbf{W}_2 ReLU(\mbf{W}_1 \mbf{F}_s^{SP}+\mbf{b}_1)+\mbf{b}_2) \in \mbb{R}^{C},
\end{equation}
where $\mbf{W}_1 \in \mbb{R}^{C\times 2C}, \mbf{b}_1 \in \mbb{R}^C, \mbf{W}_2 \in \mbb{R}^{C\times C}, \mbf{b}_2 \in \mbb{R}^C$ represent the two-layer MLP.

Finally, the signer embeddings $\mbf{E}_{sig}$ are fed into a classifier to yield signer probabilities $\mbf{p}_{sig} \in (0,1)^{N_{sig}}$, where $N_{sig}$ is the number of signers. 
The SRM is trained with the signer classification loss, which is simply a cross-entropy loss:
\begin{equation}
    \mathcal{L}_{srm} = -\log p_{sig}^i,
\end{equation}
where $i$ is the label of the signer.

\subsubsection{Gradient Reversal}
If the CSLR backbone is jointly trianed with $\mathcal{L}_{srm}$, it will become the multi-task learning, which, however, cannot promise removing the signer information from the backbone.
In this work, we treat each signer as a domain and formulate SI-CSLR as a domain generalization problem in which no test signers are seen during training. 
The gradient reversal layer was proposed in \cite{ganin2016domain} to address the domain generalization problem by learning features that are discriminative to the main classification task while indiscriminate to the domain gap.
More specifically, according to \cite{ganin2016domain}, denoting the parameters of the feature extractor, label predictor, and domain classifier as $\theta_f$, $\theta_y$, and $\theta_d$, respectively, the optimization of these parameters can be formulated as:
\begin{equation}
\label{equ:grad_rev}
\begin{split}
\theta_f &\leftarrow \text{optimizer}(\theta_f, \nabla_{\theta_f}\mathcal{L}_y, -\lambda \nabla_{\theta_f}\mathcal{L}_d, \eta), \\
\theta_y &\leftarrow \text{optimizer}(\theta_y, \nabla_{\theta_y}\mathcal{L}_y, \eta), \\
\theta_d &\leftarrow \text{optimizer}(\theta_d, \lambda \nabla_{\theta_d}\mathcal{L}_d, \eta),
\end{split}
\end{equation}
where $\mathcal{L}_y$ and $\mathcal{L}_d$ are the main classification and domain classification losses, respectively, $\lambda$ is the loss weight for $\mathcal{L}_d$, and $\eta$ is the learning rate.

We adapt Equation \ref{equ:grad_rev} by instantiating $\mathcal{L}_y$ and $\mathcal{L}_d$ as the backbone training loss $\mathcal{L}_b$ and signer classification loss $\mathcal{L}_{srm}$, which are illustrated in Figure 
\ref{fig:srm}, respectively. We also merge $\theta_f$ and $\theta_y$ as $\theta_b$ to denote the parameters of the backbone, and use $\theta_s$ to represent the parameters of the SRM. The new optimization process can be formulated as:
\begin{equation}
\begin{split}
\theta_b &\leftarrow \text{optimizer}(\theta_b, \nabla_{\theta_b}\mathcal{L}_b, -\lambda \nabla_{\theta_b}\mathcal{L}_{srm}, \eta), \\
\theta_s &\leftarrow \text{optimizer}(\theta_s, \lambda \nabla_{\theta_s}\mathcal{L}_{srm}, \eta).
\end{split}
\end{equation}

As a result, the SRM itself is trained with $\mathcal{L}_{srm}$ as usual, but the backbone is trained ``reversely" so that the extracted features cannot discriminate signers, and the signer information is implicitly removed.
We validate the effectiveness of the SRM on two challenging SI-CSLR benchmarks, establishing a strong baseline for future works on SI-CSLR.

\subsection{Alignment Module and Loss Function}
\label{sec:loss}
We follow recent works \cite{self-mutual, vac, stmc, cma} to adopt CTC-based alignment module.
It yields a label for each frame which may be a repeating label or a special blank symbol.
CTC assumes that the model output at each time step is conditionally independent of each other.
Given an input sequence $\mbf{x}$, the conditional probability of a label sequence $\boldsymbol{\phi}=\{\phi_i\}_{i=1}^T$, where $\phi_i \in \mathcal{V}\cup\{blank\}$ and $\mathcal{V}$ is the vocabulary of glosses, can be estimated by:
\begin{equation}
    p(\boldsymbol{\phi}|\mbf{x}) = \prod_{i=1}^T p(\phi_i|\mbf{x}),
\label{equ:ctc}
\end{equation}
where $p(\phi_i|\mbf{x})$ is the frame-level gloss probabilities generated by a classifier.
The final probability of the gloss label sequence is the summation of all feasible alignments:
\begin{equation}
    p(\mbf{y}|\mbf{x}) = \sum_{\boldsymbol{\phi}=\mathcal{G}^{-1}(\mbf{y})} p(\boldsymbol{\phi}|\mbf{x}),
\end{equation}
where $\mathcal{G}$ is a mapping function to remove repeats and blank symbols in $\boldsymbol{\phi}$, and $\mathcal{G}^{-1}$ is its inverse mapping.
Then the CTC loss is defined as:
\begin{equation}
    \mathcal{L}_{ctc} = -\log p(\mbf{y}|\mbf{x}).
\end{equation}
Finally, the overall loss function is a combination of the CTC, SAC, SEC, and signer classification losses:
\begin{equation}
\label{equ:overall_loss}
    \mathcal{L}=\underbrace{\mathcal{L}_{ctc}+\mathcal{L}_{sac}+\mathcal{L}_{sec}}_{\mathcal{L}_b}+\lambda\mathcal{L}_{srm},
\end{equation}
where $\lambda=0$ for signer-dependent datasets, and $\lambda>0$ for signer-independent ones.

\subsection{A Strong Sequential Module: Local Transformer}
\label{sec:lt}

\begin{figure}[t]
\begin{subfigure}[t]{.4\textwidth}
 \centering
 \includegraphics[width=\textwidth]{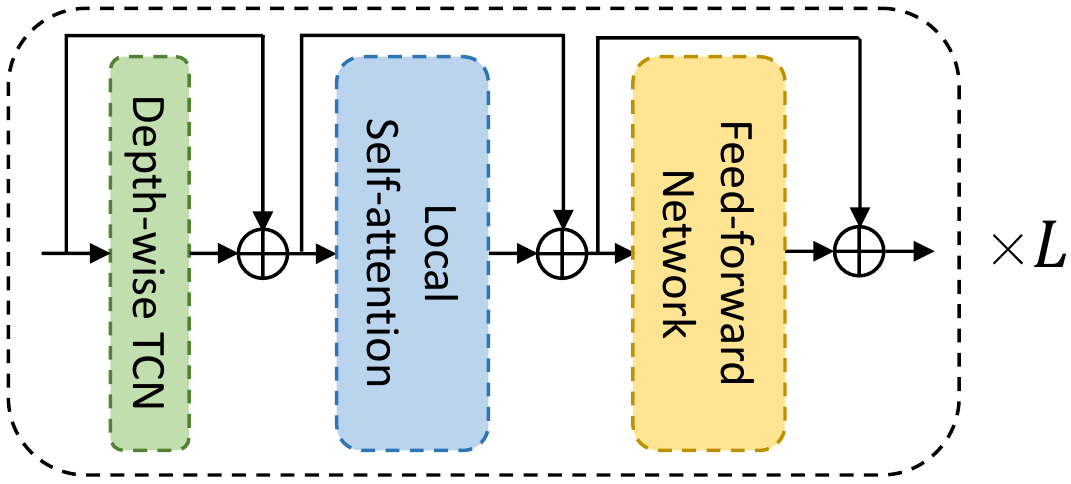}
 \caption{Local Transformer (LT). We omit LayerNorm \cite{layernorm} for simplicity.}
 \label{fig:lctr}
\end{subfigure}
\quad
\begin{subfigure}[t]{.45\textwidth}
  \centering
  \includegraphics[width=\textwidth]{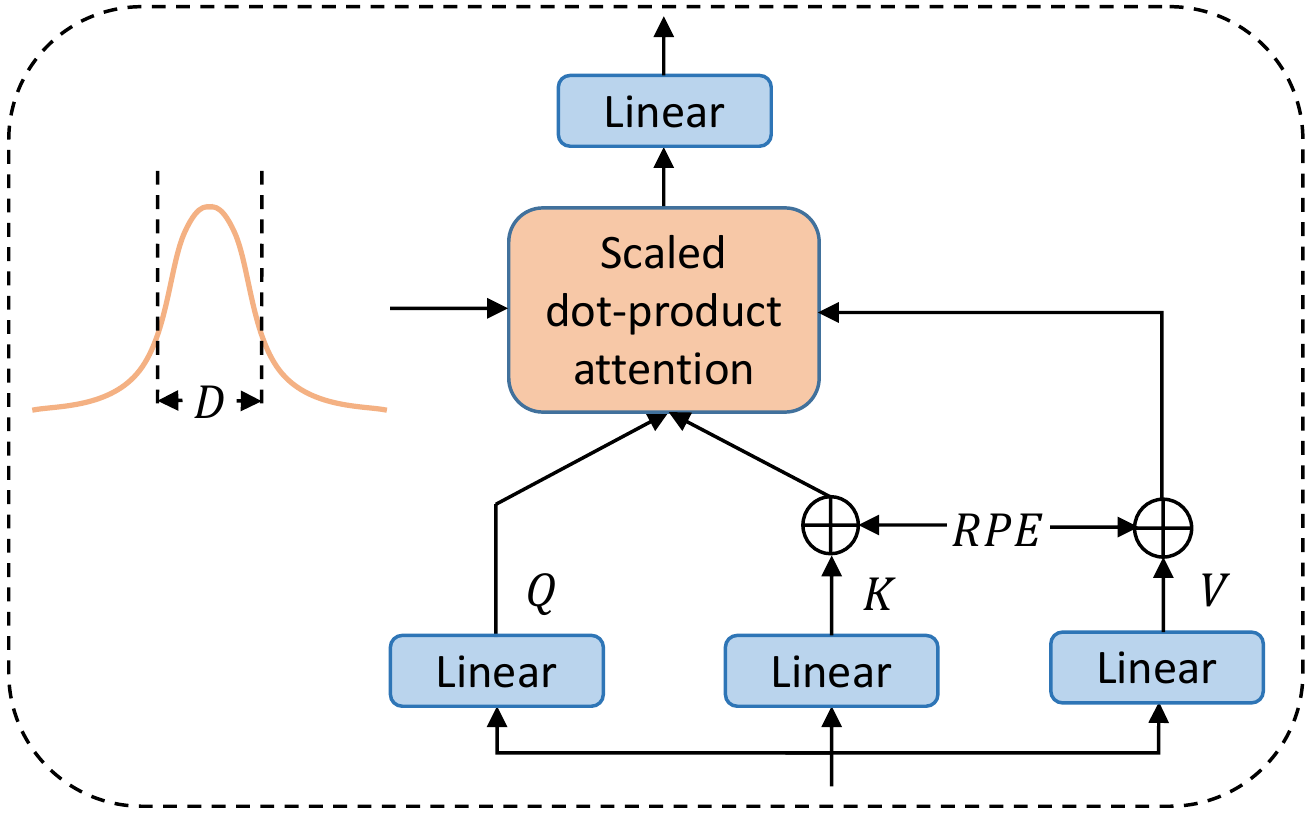}
  \caption{Local Self-attention (LSA).}
  \label{fig:lcsa}
\end{subfigure}
\caption{We propose a strong sequential module, local transformer. It is based on QANet \cite{qanet}, which validates the effectiveness of combining TCNs with self-attention. The difference is that we further leverage Gaussian bias \cite{gau-1, gau-2} to introduce local contexts into the self-attention module, \ie, local self-attention. ($L$: number of LT layers, which is set to 2 as default; $RPE$: relative positional encoding \cite{rpe}; $D$: window size of the Gaussian bias.)}
\end{figure}

The sequential module is an important component of the CSLR backbone.
Most existing CSLR works adopt globally-guided architectures, \eg, BiLSTM \cite{iopt, cma} and vanilla Transformer \cite{sfl, slt}, for sequence modeling due to their strong capability of capturing long-term temporal dependencies. 
However, within a sign video, each gloss is short, consisting of only a few frames.
This can explain why a locally-guided architecture, such as TCNs, can also achieve excellent performance \cite{fcn}.
In this subsection, we will elaborate a mixed architecture, Local Transformer (LT), which can leverage both global and local contexts for sequence modeling for CSLR.

Figure \ref{fig:lctr} shows the architecture of LT. 
Each LT layer consists of a depth-wise TCN layer, a local self-attention (LSA) layer, and a feed-forward network. 
Since the depth-wise TCN layer and the feed-forward network are the same as those used in \cite{qanet, transformer}, below we will only give the formulation of the LSA.

As shown in Figure \ref{fig:lcsa}, three linear layers first project the input feature sequence $\mathbf{A} \in \mathbb{R}^{T\times d}$ into queries $\mathbf{Q}\in \mathbb{R}^{T\times d}$, keys $\mathbf{K}\in \mathbb{R}^{T\times d}$, and values $\mathbf{V} \in \mathbb{R}^{T\times d}$, respectively.
We then split $\mbf{Q}, \mbf{K}, \mbf{V}$ into $\{\mbf{Q}^h\}_{h=1}^{N_h}, \{\mbf{K}^h\}_{h=1}^{N_h}, \{\mbf{V}^h\}_{h=1}^{N_h}$, respectively, for multi-head self-attention as \cite{transformer}, where $\mbf{Q}^h, \mbf{K}^h, \mbf{V}^h \in \mathbb{R}^{T\times d/{N_h}}$ and $N_h$ is the number of heads.
The attention scores for each head can be obtained by the scaled dot-product attention as follows:
\begin{equation}
    \mathbf{ATT} = \left\{\frac{(\mbf{Q}^h)(\mbf{K}^{h})^T}{\sqrt{d/N_h}}\right\}_{h=1}^{N_h} \in \mathbb{R}^{N_h\times T\times T}.
\end{equation}

The vanilla self-attention treats each position equally. 
To emphasize local contexts, we adopt Gaussian bias \cite{gau-1, gau-2} to weaken the interactions between distant query-key (QK) pairs.
Given a QK pair ($\mathbf{q}_i^h, \mathbf{k}_j^h$), the Gaussian bias (GB) is defined as:
\begin{equation}
    GB_{ij}^h = -\frac{(j-i)^2}{2\sigma^2},
    \label{bias}
\end{equation}
where $\sigma=\frac{D}{2}$, and $D$ is the window size of the Gaussian bias \cite{gau-1}.
Note that although we can assign Gaussian bias with a different value of $D$ for each head, we find that a common Gaussian bias among all heads suffices to boost the performance of transformer significantly.
The final attention weights for each value vector are obtained from a softmax layer, and the output of the LSA is:
\begin{equation}
\begin{cases}
    \quad \mbf{O}^h = softmax(\mbf{ATT}^h+\mbf{GB}^h)\mbf{V}^h \\
    \quad \mbf{O}^{LSA} = cat(\{\mbf{O}^h\}_{h=1}^{N_h})\mbf{W}^O \in \mathbb{R}^{T\times d} \ ,
\end{cases}
\end{equation}
where $\mbf{W}^O \in \mathbb{R}^{d\times d}$ denotes the output linear layer.

We intuitively set $D$ as the average ratio of frame length to gloss length: $D=\frac{1}{|tr|}\sum_{i=1}^{|tr|}\frac{T_i}{N_i}$,
where $|tr|$ is the number of training samples, based on the idea that a good window size should reflect the average frame length of a gloss.
More specifically, $D=6.3,15.8,5.0$ for the PHOENIX datasets, CSL, and CSL-Daily, respectively.

\section{Experiments}

\subsection{Datasets and Evaluation Metric}
\subsubsection{Datasets}
\begin{table}[t]
    \centering
    \caption{Dataset statistics.}
    \label{tab:dataset}
    \resizebox{\linewidth}{!}{
    \begin{tabular}{l|c|c|ccc|ccc|c}
    \toprule
    \multirow{2}{*}{Dataset} & \multirow{2}{*}{Language} & \multirow{2}{*}{Vocab Size} & \multicolumn{3}{c|}{\#Samples} & \multicolumn{3}{c|}{\#Signers} & Signer-\\
    & & & Train & Dev & Test & Train & Dev & Test & Independent \\
    \midrule
    PHOENIX-2014 & Germany & 1,081 & 5,672 & 540 & 629 & 9 & 9 & 9 & No \\
    PHOENIX-2014-T & Germany & 1,085 & 7,096 & 519 & 642 & 9 & 9 & 9 & No \\
    CSL-Daily & Chinese & 2,000 & 18,401 & 1,077 & 1,176 & 10 & 10 & 10 & No \\
    \midrule
    PHOENIX-2014-SI & Germany & 1,081 & 4,376 & 111 & 180 & 8 & 1 & 1 & Yes \\
    CSL & Chinese & 178 & 4,000 & N/A & 1,000 & 40 & N/A & 10 & Yes \\
    \bottomrule
    \end{tabular}
    }
\end{table}

We evaluate our method on three signer-dependent datasets (PHOENIX-2014, PHOENIX-2014-T, and CSL-Daily) and two signer-independent datasets (PHOENIX-2014-SI and CSL).
The information of these datasets, including language, vocabulary size, train/dev/test splits, number of signers, are available in Table \ref{tab:dataset}.
Compared to some widely-adopted datasets in action recognition, \eg, Kinetics-600 \cite{k600} with ~500K videos and Something-Something v2 \cite{sthsthv2} with ~169K videos, the size of these sign language datasets are quite small. This can also explain why some specific training strategies, \eg, stage optimization and auxiliary training, are suggested necessary for CSLR before.

\subsubsection{Evaluation Metric}
We use word error rate (WER) to measure the dissimilarity between two sequences.
\begin{equation}
    \text{WER} = \frac{\#\text{deletions} + \#\text{substitutions} + \#\text{insertions}}{\#\text{glosses in label}}
\end{equation}
The official evaluation scripts provided by each dataset are used for measuring the WER.

\subsection{Implementation Details}
\subsubsection{Data Augmentation}
We first resize the RGB frames to $256\times 256$ and then crop them to $224\times 224$. For the PHOENIX datasets, we adopt stochastic frame dropping (SFD) \cite{sfl} with a dropping ratio of 50\%. However, due to the longer duration of videos in CSL and CSL-Daily, we implement a \textit{seg-and-drop} strategy that first segments the videos into short clips consisting of only two frames, and then one frame is randomly dropped from each clip. By doing so, the processed videos retain half of the original frames, while preserving most information. After that, we further randomly drop 40\% frames using SFD from these processed videos.

\subsubsection{Backbones and Hyper-parameters}
We first choose three representative backbones to validate the effectiveness of our method.
\begin{itemize}
    \item VGG11+TCN+BiLSTM (VTB). It is widely adopted in some recent works \cite{stmc, vac}. VGG11 \cite{vggnet} is used as the visual module, and the sequential module is composed of the TCN and BiLSTM to capture both local and global contexts.
    
    \item CNN+TCN (CT). This lightweight backbone only consists of a 9-layer 2D-CNN and a 3-layer TCN, which is proposed in \cite{fcn}.
    
    \item VGG11+Local Transformer (VLT). The sequential module is a 2-layer local transformer encoder described in Section \ref{sec:lt}.
\end{itemize}
To better validate the robustness of our method, we additionally append our local transformer to three mainstream visual backbones including ResNet-18 \cite{resnet}, MobileNet-v3-Small \cite{mobilenet_v3}, and GoogLeNet \cite{googlenet}.
To align the channel dimensions of visual and sequential features, we configure the TCN layers in CT and VTB to have an output channel size of 512. Additionally, we set the number of hidden units for the BiLSTM in VTB to $2\times 256$. These adjustments lead to comparable Word Error Rates (WERs) to those reported in the original papers \cite{fcn, stmc}, maintaining consistency in performance evaluation.
We empirically insert the spatial attention module after the 5th CNN layer.
For post-processing, we set $\gamma_x=\gamma_y=14$ based on the experimental results presented in Section \ref{sec:gamma}.
The kernel size of the depth-wise TCN layer in both our SEE and VLT backbone models is set to 5, consistent with \cite{qanet}.
To determine the margin $\alpha$ in Equation \ref{equ:sec}, we consider the maximum difference between negative and positive cosine distances and set $\alpha$ to 2.
Regarding the signer removal module, we empirically position it after the 5th CNN layer, and the default weight for $\mathcal{L}_{srm}$, $\lambda$, is set to 0.75.

\subsubsection{Training}
All models are trained using a batch size of 2, following recent works \cite{self-mutual, vac, stmc}. We employ the Adam optimizer \cite{adam} with an initial learning rate of $1\times 10^{-4}$ and a weight decay factor of $1\times 10^{-4}$. We empirically notice that $\mathcal{L}_{sec}$ decreases at a faster rate compared to $\mathcal{L}_{ctc}$. To ensure consistent training progress between the backbone and SEE, we decrease the learning rate of the SEE by a default factor of 0.1 for the backbone architectures used in our experiments (for CT, we set the factor to 0.01).
Following the previous approach \cite{slt}, we employ a learning rate scheduling strategy known as plateau. Specifically, if the WER on the dev set does not decrease for 6 consecutive evaluation steps, the learning rate will be reduced by a factor of 0.7. However, since CSL does not have an official dev set, we decrease the learning rate after the 15th and 25th epoch, and subsequently every 5 epochs after the 30th epoch. The total number of training epochs is set to 60.

\subsubsection{Inference and Decoding.}
Following \cite{sfl}, to match the training condition, we evenly select every $\frac{1}{p_d}$-th frame to drop during inference, where $p_d$ is the dropping ratio.
We adopt the beam search algorithm with a beam size of 10 for decoding.

\begin{table}[t]
  \centering
  \caption{Ablation study for SAC and SEC. During inference, since our SEC can be removed, only the spatial attention module in SAC will introduce negligible parameters and affect inference speed. ($\text{SAC}^-$ denotes only inserting the spatial attention module but not guided by $\mathcal{L}_{sac}$; Par.: number of parameters; Sp.: inference speed measured on the same TITAN RTX GPU in seconds per video.)}
  \resizebox{\linewidth}{!}{
  \begin{tabular}{c|ccc|ccc|c|ccc|ccc}
    \toprule
    Backbone & $\text{SAC}^-$ & SAC & SEC & WER\% & Par.(M) & Sp.(s) & Backbone & $\text{SAC}^-$ & SAC & SEC & WER\% & Par.(M) & Sp.(s)\\
    \midrule
    \multirow{5}{*}{VTB} & & & & 25.0 & 15.6359 & 0.169 & \multirow{5}{*}{\shortstack{ResNet-18\\+LT}} & & & & 23.8 & 18.1356 & 0.103\\
    & \checkmark & & & 24.6 & +0.0001 & +0.002 & & \checkmark & & & 23.8 & +0.0001 & +0.002\\
    & & \checkmark & & 23.7 & +0.0001 & +0.002 & & & \checkmark & & 22.6 & +0.0001 & +0.002\\
    & & & \checkmark & 24.3 & +0.0000 & +0.000 & & & & \checkmark & 22.8 & +0.0000 & +0.000\\
    & & \checkmark & \checkmark & \tbf{22.6} & +0.0001 & +0.002 & & & \checkmark & \checkmark & \tbf{22.2} & +0.0001 & +0.002\\
    
    \midrule
    \multirow{5}{*}{CT} & & & & 26.1 & 8.7504 & 0.095 & \multirow{5}{*}{\shortstack{MobileNet-v3-Small\\+LT}} & & & & 26.0 & 9.0502 & 0.098\\
    & \checkmark & & & 26.0 & +0.0001 & +0.001 & & \checkmark & & & 25.8 & +0.0001 & +0.001\\
    & & \checkmark & & 25.1 & +0.0001 & +0.001 & & & \checkmark & & 25.2 & +0.0001 & +0.001\\
    & & & \checkmark & 25.2 & +0.0000 & +0.000 & & & & \checkmark & 25.2 & +0.0000 & +0.000\\
    & & \checkmark & \checkmark & \tbf{24.5} & +0.0001 & +0.001 & & & \checkmark & \checkmark & \tbf{24.7} & +0.0001 & +0.001\\
    
    \midrule
    \multirow{5}{*}{VLT} & & & & 21.5 & 16.1850 & 0.163 & \multirow{5}{*}{\shortstack{GoogLeNet\\+LT}} & & & & 24.0 & 23.7070 & 0.112 \\
    & \checkmark & & & 21.4 & +0.0001 & +0.002 & & \checkmark & & & 23.9 & +0.0001 & +0.002\\
    & & \checkmark & & 20.8 & +0.0001 & +0.002 & & & \checkmark & & 23.4 & +0.0001 & +0.002\\
    & & & \checkmark & 20.9 & +0.0000 & +0.000 & & & & \checkmark & 23.3 & +0.0000 & +0.000\\
    & & \checkmark & \checkmark & \tbf{20.4} & +0.0001 & +0.002 & & & \checkmark & \checkmark & \tbf{22.9} & +0.0001 & +0.002\\
    \bottomrule
  \end{tabular}
  }
  \label{tab:sacsec}
\end{table}

\begin{figure}[t]
  \centering
   \includegraphics[width=1.0\linewidth]{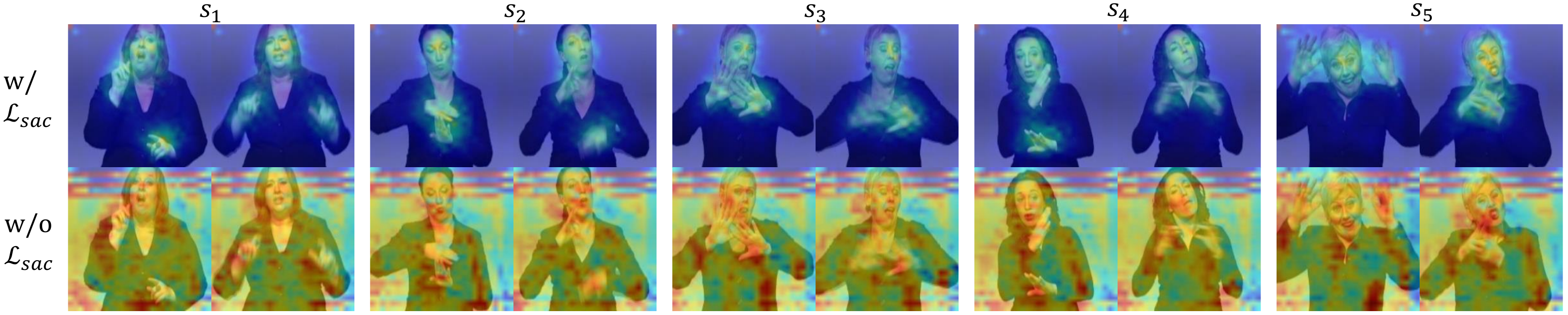}
   \caption{Visualization results for learned spatial attention masks with or without the guidance of $\mathcal{L}_{sac}$. We randomly select five samples ($s_1, \dots, s_5$) from the \tbf{test} set, and for each sample, we select one clear frame and one blurry frame. It is clear that the guidance of $\mathcal{L}_{sac}$ can help the spatial attention module capture the informative regions (face and hands) more accurately.}
   \label{fig:vis_sac}
\end{figure}

\subsection{Ablation Studies for $\text{C}^2$SLR}
We first conduct ablation studies for $\text{C}^2$SLR on PHOENIX-2014 following previous works \cite{vac, stmc, cma, self-mutual}.

\subsubsection{Effectiveness of SAC and SEC}
As shown in Table \ref{tab:sacsec}, both SAC and SEC generalize well on different backbones: the performance of all the six backbones can be clearly improved.
However, if the spatial attention module is inserted into the backbones without any guidance, \ie, $\text{SAC}^-$, the model performance can only be improved slightly, which verifies the effectiveness of $\mathcal{L}_{sac}$.
The effectiveness of SEC suggests that explicitly enforcing the consistency between the visual and sequential modules at the sentence level can strengthen the cross-module cooperation, which leads to the performance gain.
The improvements due to SAC and SEC are complementary so that using both of them can obtain better results than using only one of them.
Besides, since VLT performs the best among the six backbones, we will use it as the default backbone for the following experiments.

\subsubsection{Visualization Results for SAC}
Figure \ref{fig:vis_sac} shows the visualization results of the learned spatial attention masks of SAC (with $\mathcal{L}_{sac}$) and $\text{SAC}^-$ (without $\mathcal{L}_{sac}$) for five test samples.
It should be noted that since SAC is deactivated during testing, our comparison is fair.
First, it is quite clear that the learned attention masks with the guidance of $\mathcal{L}_{sac}$ look much better.
Without the guidance of $\mathcal{L}_{sac}$, the attention masks are quite messy with horizontal lines at the top and many highlights at trivial regions, \eg, the left shoulder of $s_2$, the hair of $s_1$ and $s_4$, and the waist of $s_3, s_5$.
This explains why $\text{SAC}^-$ can only slightly improve the performance of the backbones as shown in Table \ref{tab:sacsec}.
Second, our SAC is so robust that the IRs (face and hands) in blurry frames (right columns of $s_1$ to $s_5$) can still be captured precisely.
Third, it is capable of dealing with different hand positions: \eg, both two hands are lower than the face ($s_1, s_3$); one hand is near the face while the other one is not ($s_1, s_2, s_4$), and hands are overlapped ($s_5$).

\begin{wraptable}{r}{0.5\textwidth}
  \caption{Ablation study for SAC.}
  \resizebox{\linewidth}{!}{
  \begin{tabular}{l|cc}
    \toprule
    Method & WER\% & \#Param(M)\\
    \midrule
    VLT + SAC & \tbf{20.8} & 16.1851 \\
    \ \ - channel weights & 21.3 & -0.0000 \\
    \ \ \ \ + channel attention  \cite{woo2018cbam} & 21.2 & +0.0335\\
    \ \ - post-processing & 21.7 & -0.0000 \\
    \ \ - face & 21.1 & -0.0000 \\
    \ \ - hands & 21.2 & -0.0000 \\
    \bottomrule
  \end{tabular}
  }
  \label{tab:sac}
\end{wraptable}

\subsubsection{Channel Weights}
Within our spatial attention module, each channel can receive a weight to better measure its importance before squeezing the feature maps.
Removing the channel weights degenerates to the channel-wise average pooling in CBAM \cite{woo2018cbam} and achieves a WER of 21.3\%, which leads to a performance drop by 0.5\% as shown in Table \ref{tab:sac}.
Although our channel weights share a similar idea with the channel attention module of CBAM, which builds extra linear layers to generate the attention weights, no extra parameters are introduced in our spatial attention module.
To further validate their effectiveness, after removing the channel weights, we conduct one more experiment by adding the channel attention module back as CBAM; however, it can only lead to a slight performance gain and cannot outperform ours even with extra parameters.

\subsubsection{Heatmap Refinement}
We discuss in the Section \ref{sec:post-process} that the raw heatmaps of HRNet \cite{sun2019deep} consist of too many defects which may hinder the learning of the spatial attention module.
As shown in Table \ref{tab:sac}, the quality of the keypoints heatmaps can make a difference on model performance: directly using the original heatmaps without post-processing achieves a WER of 21.7\%, which reduces the performance of SAC by almost 1\%.

\subsubsection{Effect of Each Informative Region}
As shown in the last two rows in Table \ref{tab:sac}, removing either face or hands region can harm the performance of SAC. The results validate that both signers' face and hands play a key role in conveying information, which is also mentioned in \cite{stmc, koller2020quantitative}.

\begin{figure}[!t]
\centering
\includegraphics[width=0.7\linewidth]{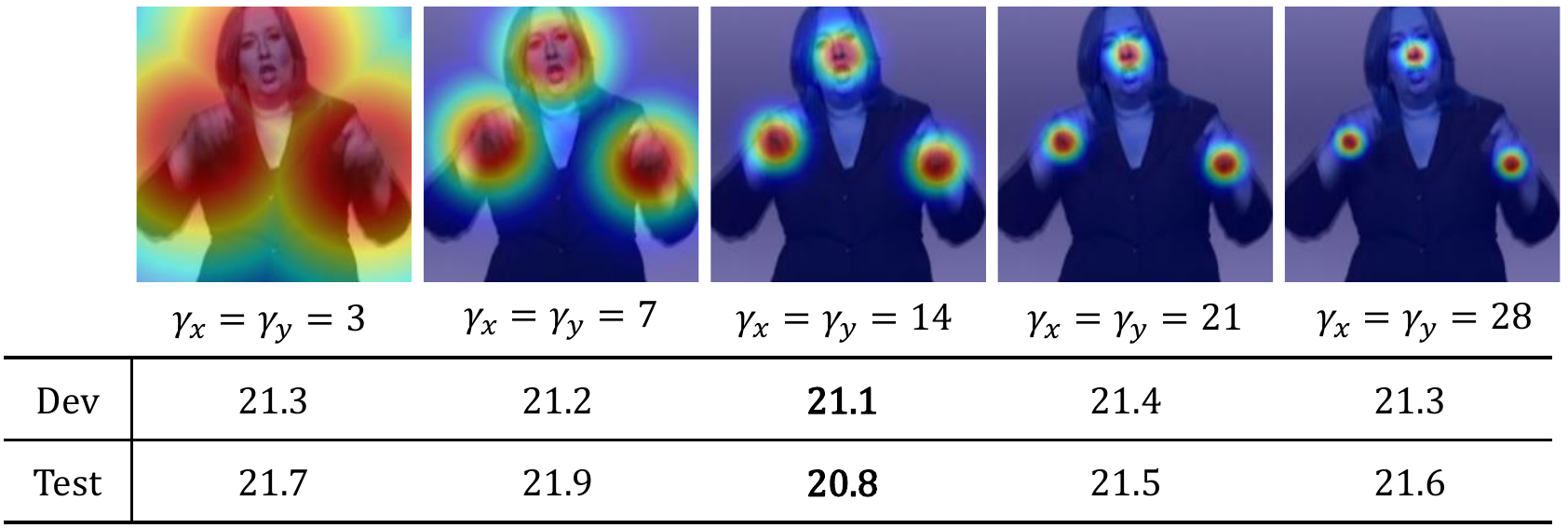}
\caption{Visualization results and performance comparison for different $\gamma_x, \gamma_y$ in Equation \ref{equ:post}. Since for real practice, the height and the width of the spatial attention masks are usually the same, we set $\gamma_x$ and $\gamma_y$ to the same value.}
\label{fig:gamma}
\end{figure}

\subsubsection{Effect of the Hyper-parameters $\gamma_x, \gamma_y$ of Equation \ref{equ:post}}
\label{sec:gamma}
We think $\gamma_x$ and $\gamma_y$ are two important hyper-parameters since they control the scale of highlighted regions in keypoints heatmaps.
Thus, we conduct experiments to compare the performance of different $\gamma_x, \gamma_y$ as shown in Figure \ref{fig:gamma}. 
The model performance is worse when they are either too large (cannot cover the informative regions entirely) or too small (cover too many trivial regions).
When $\gamma_x=\gamma_y=14$, the model achieves the best performance.

\begin{table*}[t]
\caption{Ablation study for SEC.}
\begin{subtable}[t]{0.48\textwidth}
\caption{Ablation study for the architecture of the sentence embedding extractor and negative sampling. (TF: Transformer; DTCN: depth-wise TCN; Neg. Sam.: negative sampling.)}
\centering
\resizebox{\linewidth}{!}{
\begin{tabular}{l|cc|c}
    \toprule
    Method & Extractor & Neg. Sam. & WER\% \\
    \midrule
    \multirow{4}{*}{VLT + SEC} & TF+DTCN & \checkmark & \tbf{20.9} \\
    & TF+DTCN & \texttimes & 21.5 \\
    & TF & \checkmark & 21.1 \\
    & BiLSTM & \checkmark & 21.3 \\
    \bottomrule
\end{tabular}
}
\label{tab:sec_ext}
\end{subtable}
\quad
\begin{subtable}[t]{0.48\textwidth}
\caption{Ablation study for the constraint level. We fine-tune the loss factor of VA as \cite{vac} on the VLT for fair comparisons.}
\centering
\resizebox{\linewidth}{!}{
\begin{tabular}{l|c|c}
    \toprule
    Level & Constraint & WER\% \\
    \midrule
    Sentence & consistency & \tbf{20.9} \\
    \midrule
    \multirow{4}{*}{Frame} & consistency & 21.6 \\
    & visual enhancement (VE) \cite{vac} & 22.3 \\
    & visual alignment (VA) \cite{vac} & 21.9 \\
    & VE+VA \cite{vac} & 22.8 \\
    \bottomrule
\end{tabular}
}
\label{tab:sec_lev}
\end{subtable}
\label{tab:sec}
\end{table*}
\subsubsection{Sentence Embedding Extractor and Negative Sampling}
Our sentence embedding extractor consists of a depth-wise TCN layer and a transformer encoder aiming to model local and global contexts, respectively.
As shown in Table \ref{tab:sec_ext}, local contexts are important to sentence embedding extraction as dropping the TCN layer would lead to worse performance.
We also compare our method with the common practice, which concatenates the last two hidden states of BiLSTM and treats it as the sentence embedding.
Nevertheless, that it underperforms the transformer-based extractors implies the strength of the self-attention mechanism for sentence embedding extraction.
Table \ref{tab:sec_ext} also shows that negative sampling plays a key role in our SEC: without negative sampling, that is, directly minimizing the sentence embedding distance between the visual and sequential features, is not effective.

\subsubsection{Constraint Level}
\label{sec:cons_lev}
As shown in Table \ref{tab:sec_lev}, we implement some frame-level constraints to validate the effectiveness of our SEC.
First, we replace the sentence embeddings, $\mbf{v}_{se}$ and $\mbf{s}_{se}$ in Equation \ref{equ:sec}, by their corresponding frame-level features to minimize the positive distances while maximizing the negative distances at the frame level.
However, it leads to a performance degradation of 0.7\% compared to our SEC.
We further compare our SEC with VAC \cite{vac}, which is composed of two frame-level constraints: visual enhancement (VE) and visual alignment (VA).
First, an extra classifier is appended to the visual module to yield frame-level probability distributions (visual distribution).
VE is implemented as a CTC loss computed between the visual distribution and the gloss label, which is the same as the one used for training the backbone.
Second, VA is simply a KL-divergence loss, which aims to minimize the distance between the visual distribution and the original probability distribution ($p(\phi_i|\mbf{x})$ in Equation \ref{equ:ctc}).
Table \ref{tab:sec_lev} shows that both VE and VA perform much worse than our SEC.
The results suggest that our SEC is a more proper way to measure the consistency between the visual and sequential modules.

\begin{figure}[t]
	\centering
	\includegraphics[width=0.7\linewidth]{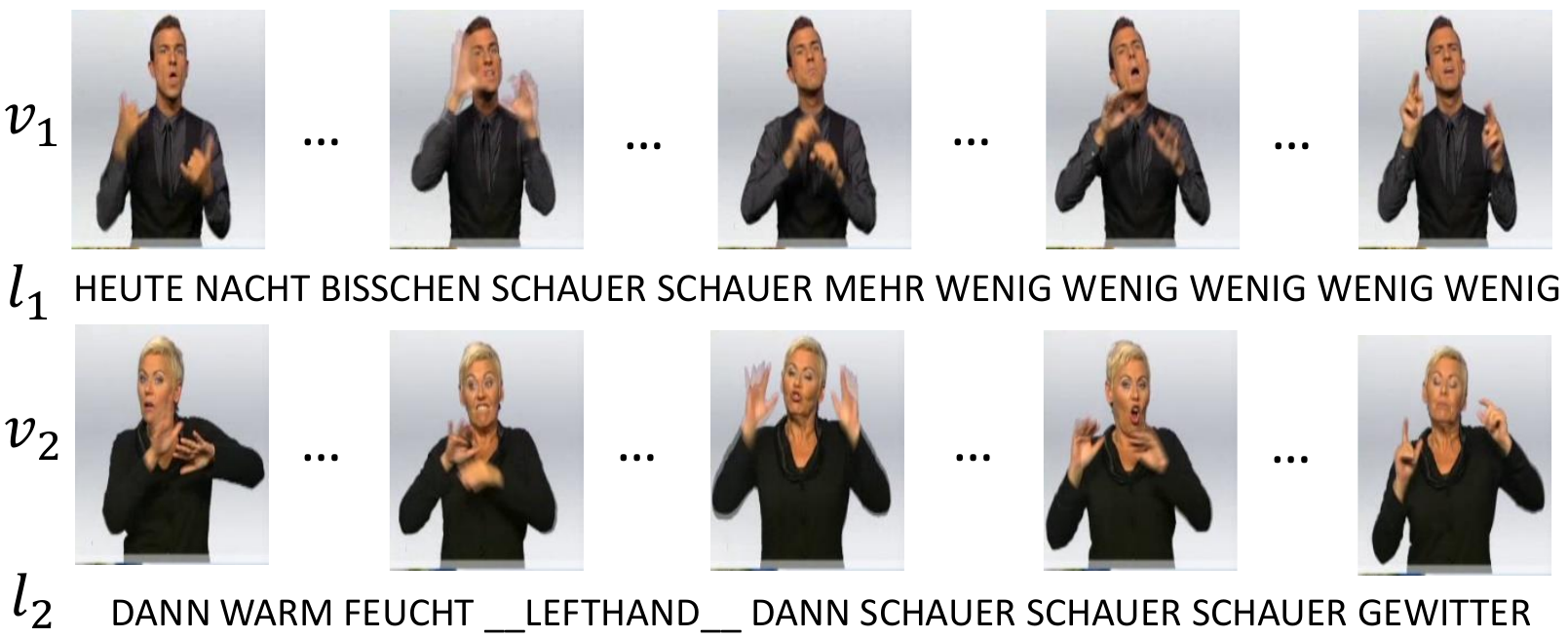}
	\caption{Two examples of video-gloss pairs. For $i\in\{1,2\}$, $v_i$ denotes the video, and for $i\in\{1,2\}$, $l_i$ denotes the corresponding gloss annotations. Their sentence embedding distances are shown in Table \ref{tab:vgpair}.}
	\label{fig:vgpair}
\end{figure}

\begin{table}[t]
    \begin{minipage}{.4\linewidth}
    \caption{Examples of sentence embedding distances of the visual and sequential features. $v_1$ and $v_2$ are the videos in Figure \ref{fig:vgpair}.}
    \centering
    \begin{tabular}{c|cc}
    	\toprule
        $d(\cdot,\cdot)$ & $\mbf{E}_{sen}^s(v_1)$ & $\mbf{E}_{sen}^s(v_2)$ \\
        \midrule
        $\mbf{E}_{sen}^v(v_1)$ & 0.01 & 1.99 \\
        $\mbf{E}_{sen}^v(v_2)$ & 1.76 & 0.37 \\
        \bottomrule
    \end{tabular}
    \label{tab:vgpair}
    \end{minipage}%
    \hspace{8pt}
    \begin{minipage}{.55\linewidth}
    \centering
    \vspace{10pt}
    \caption{Effect of the value of $\lambda$ (the weight for the loss $\mathcal{L}_{srm}$) in Equation \ref{equ:overall_loss}.}
    \begin{tabular}{l|ccccccc}
        \toprule
        $\lambda$ & 0 & 0.25 & 0.5 & 0.75 & 1.0 & 1.25 & 1.5 \\
        \midrule
        Dev & 34.3 & 35.1 & 35.3 & \textbf{33.1} & 33.5 & 35.0 & 34.4 \\
        Test & 34.4 & 33.8 & 33.1 & \textbf{32.7} & 32.8 & 34.2 & 33.6 \\
        \bottomrule
    \end{tabular}
    \label{tab:lambda}
    \end{minipage}
\end{table}

\subsubsection{Examples of Video-gloss Pairs}
To verify whether $\mathcal{L}_{sec}$ can really separate positive and negative samples, we provide two examples of video-gloss pairs denoted as $(v_1, l_1)$ and $(v_2, l_2)$ as shown in Figure \ref{fig:vgpair}.
The sentence embedding distances of the visual and sequential features of $v_1$ and $v_2$ are shown in Table \ref{tab:vgpair}. 
It is clear that the distance between the two features of the same video (diagonal entries, positive pairs) can be very small. 
Otherwise (off-diagonal entries, negative pairs), the distance can be very large (the maximum value is 2.00).

\subsection{Ablation Studies for the Signer Removal Module}
We further conduct ablation studies for our signer removal module (SRM) on the challenging signer-independent dataset, PHOENIX-2014-SI.

\subsubsection{Effect of the Hyper-parameter $\lambda$ of Equation \ref{equ:overall_loss}}
\label{sec:lambda}
According to \cite{liu2018exploring}, the weight for the domain classification loss, \ie, our signer classification loss $\mathcal{L}_{srm}$, is an important hyper-parameter.
We fine-tune it from 0 to 1.5 with an interval of 0.25 as shown in Table \ref{tab:lambda}.
When $\lambda=0$, the model degenerates to $\text{C}^2$SLR and performs worse than other models with $\lambda>0$.
The results suggest the importance of removing signer information for SI-CSLR.
When $\lambda=0.75$, the model can achieve the best performance with a WER of 33.1\%/32.7\% on the dev and test set, respectively.

\begin{table}[t]
  \centering
  \caption{Ablation study for the signer removal module. Experiments are conducted on PHOENIX-2014-SI. (SP: statistics pooling; GR: gradient reversal)}
  \begin{tabular}{l|ccc|c|c}
    \toprule
    Method & $\mathcal{L}_{srm}$ & SP & GR & WER\% & Type \\
    \midrule
    \multirow{6}{*}{$\text{C}^2$SLR+} & & & & 34.4 & N/A \\
    \cmidrule(){2-6}
    & \checkmark & & & 34.9 & \multirow{2}{*}{Multi-task Learning} \\
    & \checkmark & \checkmark & & 33.5 & \\
    \cmidrule(){2-6}
    & \checkmark & & \checkmark & 33.6 & \multirow{2}{*}{Feature Disentanglement} \\
    & \checkmark & \checkmark & \checkmark & \textbf{32.7} & \\
    \bottomrule
  \end{tabular}
  \label{tab:srm}
\end{table}
\subsubsection{Statistics Pooling and Gradient Reversal}
We further conduct ablation studies for the two major components of our SRM, the statistics pooling (SP) and gradient reversal (GR) layer.
First, the use of the GR layer decides the type of learning method: feature disentanglement or multi-task learning.
As shown in Table \ref{tab:srm}, it is clear that with the use of GR, models under the feature disentanglement setting can significantly outperform those under the multi-task learning setting.
The result implies that removing signer information is effective to SI-CSLR.
However, we find that the model, $\text{C}^2$SLR+SP, can also outperform the baseline under the multi-task setting.
We think this is because the multi-task learning can be seen as a kind of regularization \cite{zhang2021survey}, which endows the shared networks between the CSLR and signer classification branches with better generalization capability.
Similar ideas also appear in some works that jointly train a speech recognition model and a speaker recognition model \cite{liu2018speaker, pironkov2016speaker}.
Finally, the effectiveness of SP also validates that sentence-level signer embeddings are more robust than frame-level ones to achieve signer classification, leading to better performance.

\begin{table}[t]
\centering
\caption{Performance comparison between seen and unseen signers.}
\begin{tabular}{l|c|c|c}
\toprule
\multirow{2}{*}{Method} & Seen Signers & Unseen Signers & Relative Gap \\
& (WER\%) & (WER\%) & (\%) \\
\midrule
$\text{C}^2$SLR & 22.7 & 34.4 & 51.5 \\
$\text{C}^2$SLR + SRM & 23.0 & 32.7 & \textbf{42.2} \\
\bottomrule
\end{tabular}
\label{tab:srm_gap}
\end{table}
\subsubsection{Effect of the SRM over Seen and Unseen Signers}
We finally study the effect of the SRM over seen and unseen signers. We first build an extra test set consisting of only seen signers during training by removing videos performed by unseen signers from the official test set of PHOENIX-2014, and then retest ``$\text{C}^2$SLR" and ``$\text{C}^2$SLR+SRM" on this extra test set. As shown in Table \ref{tab:srm_gap}, with a comparable performance over the seen signers, adding the SRM can significantly narrow the performance gap between unseen and seen signers. The results suggest that our SRM can be more helpful for the real-world situation that most testing signers are unseen.

\subsection{Comparison with State-of-the-art Results}
\begin{table*}[!t]
\centering
\caption{Comparison on signer-dependent datasets. (R: RGB; F: optical flow; P: pose.)}
\resizebox{\linewidth}{!}{
\begin{tabular}{l|c|cc|cc|cc|cc}
\toprule
\multirow{2}{*}{Method}& \multirow{2}{*}{End-to-end} & \multicolumn{2}{c|}{Modalities} & \multicolumn{2}{c|}{PHOENIX-2014} & \multicolumn{2}{c|}{PHOENIX-2014-T} & \multicolumn{2}{c}{CSL-Daily} \\
& & Training & Inference & Dev & Test & Dev & Test & Dev & Test \\
\midrule
CNN-LSTM-HMMs \cite{cnn-lstm-hmm} & \texttimes & R & R & 26.0 & 26.0  & 22.1 & 24.1 & -- & -- \\
DNF (RGB) \cite{dnf} + SBD-RL \cite{wei2020semantic} & \texttimes & R & R & 23.4 & 23.5 & -- & -- & -- & -- \\
DNF \cite{dnf} & \texttimes & R+F & R+F & 23.1 & 22.9 & -- & -- & 32.8 & 32.4 \\
CMA \cite{cma} & \texttimes & R & R & 21.3 & 21.9 & -- & -- & -- & -- \\
SMKD \cite{self-mutual} & \texttimes & R & R & 20.8 & 21.0 & 20.8 & 22.4 & -- & -- \\
STMC \cite{stmc} & \texttimes & R+P & R & 21.1 & 20.7 & 19.6 & 21.0 & -- & -- \\
\midrule
LS-HAN \cite{csl-2} & \checkmark & R & R & -- & 38.3 & -- & -- & 39.0 & 39.4 \\
TIN + Transformer \cite{zhou2021improving} & \checkmark & R & R & -- & -- & -- & -- & 33.6 & 33.1 \\
SFL \cite{sfl} & \checkmark & R & R & 24.9 & 25.3 & 25.1 & 26.1 & -- & -- \\
FCN \cite{fcn} & \checkmark & R & R & 23.7 & 23.9 & 23.3 & 25.1 & 33.2 & 32.5 \\
LCSA \cite{zuo22_interspeech} & \checkmark & R & R & 21.4 & 21.9 & --& -- & -- & --\\
SLT \cite{slt} & \checkmark & R & R & -- & -- & 24.6 & 24.5 & 33.1 & 32.0 \\
VAC \cite{vac} & \checkmark  & R & R & 21.2 & 22.3 & -- & -- & -- & -- \\
MMTLB \cite{chen2022simple} & \checkmark & R & R & -- & -- & 21.9 & 22.5 & -- & -- \\
$\text{C}^2$SLR (ours) & \checkmark & R+P & R & \tbf{20.5} & \tbf{20.4} & 20.2 & \tbf{20.4} & \tbf{31.9} & \tbf{31.0} \\
\bottomrule
\end{tabular}
}
\label{tab:SD}
\end{table*}
\subsubsection{Signer-dependent}
As shown in Table \ref{tab:SD}, we first evaluate our $\text{C}^2$SLR on three signer-dependent benchmarks: PHOENIX-2014, PHOENIX-2014-T, and CSL-Daily.

Our $\text{C}^2$SLR follows the idea of auxiliary learning, which also appears in some existing works, \eg, FCN \cite{fcn} and VAC \cite{vac}.
FCN proposes a gloss feature enhancement (GFE) module to introduce auxiliary supervision signals into the model training process.
However, the GFE module highly relies on pseudo labels (CTC decoded results), which may contain too many errors.
Our method only relies on pre-extracted heatmaps, which are quite accurate with the help of our post-processing algorithm, and the model's inherent consistency: the visual and sequential features represent the same sentence.
These two properties enable our method to outperform FCN by more than 3\% on both PHOENIX-2014 and PHOENIX-2014-T.
Recently, VAC proposes two auxiliary losses at the frame-level, which are not quite appropriate and perform worse than ours according to the comparison in Section \ref{sec:cons_lev}.
The SOTA work, STMC \cite{stmc}, adopts the complicated stage optimization strategy, which introduces extra hyper-parameters, and needs to manually decide when to switch to a new stage.
Our method is totally end-to-end trainable, and it can outperform STMC on both PHOENIX-2014 and PHOENIX-2014-T.
To the best of our knowledge, this is the first time that an end-to-end method can outperform those using the stage optimization strategy.

In terms of modality usage, our method just uses extra pose modality during training, while only RGB videos are needed for inference.
Thus, it is simpler for real application compared to DNF \cite{dnf} which is built on a two-stream architecture taking both RGB videos and optical flow as inputs.

Finally, the results reported on CSL-Daily may be more important due to its large vocabulary size.
Our method can still achieves SOTA performance on this large-scale dataset, which also validates the generalization capability of our method over different sign languages.

\begin{table*}[t]
\centering
\caption{Comparison on signer-independent datasets. (R: RGB; F: optical flow; P: pose; D: depth.)}
\begin{subtable}[t]{1.0\textwidth}
\centering
\caption{PHOENIX-2014-SI.}
\begin{tabular}{l|c|cc|cc}
    \toprule
    \multirow{2}{*}{Method} & \multirow{2}{*}{End-to-end} & \multicolumn{2}{c|}{Modalities} & \multirow{2}{*}{Dev} & \multirow{2}{*}{Test} \\
    & & Training & Inference & & \\
    \midrule
    Re-sign \cite{re-sign} & \texttimes & R & R & 45.1 & 44.1 \\
    DNF \cite{dnf} & \texttimes & R+F & R+F & 36.0 & 35.7 \\
    CMA \cite{cma} & \texttimes & R & R & 34.8 & 34.3 \\
    \midrule
    $\text{C}^2$SLR (ours) & \checkmark & R+P & R & 34.3 & 34.4 \\
    $\text{C}^2$SLR + SRM (ours) & \checkmark & R+P & R & \textbf{33.1} & \textbf{32.7} \\
    \bottomrule
\end{tabular}
\label{tab:2014SI}
\end{subtable}

\begin{subtable}[t]{1.0\textwidth}
\centering
\caption{CSL.}
\begin{tabular}{l|c|cc|c}
    \toprule
    \multirow{2}{*}{Method} & \multirow{2}{*}{End-to-end} & \multicolumn{2}{c|}{Modalities} & \multirow{2}{*}{Test} \\
    & & Training & Inference & \\
    \midrule
    LS-HAN \cite{csl-2} & \texttimes & R & R & 17.3 \\
    DPD + TEM \cite{csl-3} & \texttimes & R & R & 4.7 \\
    STMC \cite{stmc} & \texttimes & R+P & R & 2.1 \\
    \midrule
    CTF \cite{ctf} & \checkmark & R & R & 11.2 \\
    HLSTM-attn \cite{HLSTM-attn} & \checkmark & R & R & 10.2 \\
    FCN \cite{fcn} & \checkmark & R & R & 3.0 \\
    VAC \cite{vac} & \checkmark & R & R & 1.6 \\
    MSeqGraph \cite{tang2021graph} & \checkmark & R+P+D & R+P+D & 0.6 \\
    $\text{C}^2$SLR (ours) & \checkmark & R+P & R & 0.90 \\
    $\text{C}^2$SLR + SRM (ours) & \checkmark & R+P & R & 0.68 \\
    \bottomrule
\end{tabular}
\label{tab:csl}
\end{subtable}
\label{tab:SI}
\end{table*}

\subsubsection{Signer-independent}
As shown in Table \ref{tab:SI}, we further evaluate our SRM on the following two signer-independent benchmarks: PHOENIX-2014-SI and CSL.

Although some works, \eg, DNF \cite{dnf} and CMA \cite{cma}, evaluate their method on PHOENIX-2014-SI, none of them propose any dedicated module to deal with the challenging SI setting.
In this work, we develop a simple yet effective signer removal module (SRM) for SI-CSLR to make the model more robust to signer discrepancy. 
As shown in Table \ref{tab:2014SI}, our $\text{C}^2$SLR can already achieve competitive performance on PHOENIX-2014-SI, and the SRM can further improve the performance significantly.
The result validates that feature disentanglement is an effective method to remove signer-relevant information, and we believe our SRM can serve as a strong baseline for future works on SI-CSLR.

As shown in Table \ref{tab:csl}, our SRM can lead to a relative performance gain of 24.4\% over the baseline $\text{C}^2$SLR on CSL\footnote{Although the SI setting itself is challenging, since the sentences in the CSL test set all appear in the training stage, the WER can be very low ($<1$\%).}.
It is worth noting that the SOTA work, MSeqGraph \cite{tang2021graph}, uses three modalities including RGB, pose, and depth.
However, our method only uses RGB and pose information for training, and only RGB frames are needed for inference.
Thus, with a comparable performance to the SOTA work, we believe our method is more applicable in real practice.

\section{Conclusion and Future Works}
In this work, we propose three auxiliary tasks to enhance CSLR backbones. The first task requires the model to learn informative attention maps from a keypoint-guided spatial attention module. The second task enhances the representation power of visual and sequential features by imposing a sentence embedding consistency constraint. Finally, the third task enforces the model to dispel signer information with a dedicated signer removal module for the signer-independent setting. We conduct sufficient ablation studies to validate the effectiveness of the three auxiliary tasks. Remarkably, our model can achieve SOTA or competitive performance on five benchmarks, while the whole model is trained in an end-to-end manner.

Below are some directions which deserve attention for future works. First, to enhance the quality of keypoints heatmaps, lightweight keypoints estimators which can be co-trained with the CSLR backbone are necessary. Second, more advanced cross-modality sentence embedding extractor shall be considered. Finally, we believe more attention should be paid on signer-independent CSLR since it is more realistic than its signer-dependent counterpart.

\begin{acks}
The work described in this paper was supported by a grant from the Research Grants Council of the Hong Kong Special Administrative Region, China (Project No. HKUST16200118).
\end{acks}

\bibliographystyle{ACM-Reference-Format}
\bibliography{main}

\appendix
\section{Appendix}
\subsection{A Sample for Word Error Rate}
Word error rate (WER) is a widely adopted evaluation metric to meaure the dissimilarity between two sequences, which is commonly used in speech recognition and sign language recognition systems \cite{2014,2014T,cma,stmc,vac,self-mutual}. It is defined as the ratio of the number of errors to the number of words (glosses) in the transcription after aligning the prediction sequence with the label sequence:
\begin{equation}
    \text{WER} = \frac{\#\text{deletions} + \#\text{substitutions} + \#\text{insertions}}{\#\text{glosses in label}},
\end{equation}
where \# denotes "the number of". Table \ref{tab:wer} shows an example with a WER of 50\%:
\begin{table}[ht]
    \centering
    \caption{An example of WER computing. "A","B","C", and "D" represent words (glosses). Substitutions and insertions are highlighted in \textcolor{red}{red} and \textcolor{green}{green}, respectively. $\square$ denotes deletions.}
    \label{tab:wer}
    \begin{tabular}{l|l}
    \toprule
         Prediction & A\ \textcolor{red}{C}\ $\square$ C\ A\ D\ \textcolor{green}{B} \\
         Label & A\ A\ B\ C\ A\ D \\
    \bottomrule
    \end{tabular}
\end{table}

\subsection{Loss Normalization}
As shown in Table \ref{tab:loss_norm}, we conduct another experiment on Phoenix-2014 by normalizing each loss term ($\mathcal{L}_{ctc}, \mathcal{L}_{sac}, \mathcal{L}_{sec}$) of $\mathcal{L}_b$ into the range of [0,1]. Note that since $\mathcal{L}_{ctc}$ is defined as the negative log-likelihood of all feasible alignment paths, we use the reciprocal of its maximum training loss value (about 1/7) as its weight for normalization. $\mathcal{L}_{sac}$ is defined as an MSE loss between attention maps and keypoint heatmaps, which ranges from 0 to 1, and thus we set its weight to 1. $\mathcal{L}_{sec}$ is defined as a triplet loss: $\mathcal{L}_{sec}=\max\{d(x,x_p)-d(x,x_n)+\alpha, 0\}$, where $d(x,x_p)=1-\frac{x\cdot x_p}{\|x\|_2\|x_p\|_2} \in [0,2]$ and $\alpha=2$. Thus, the theoretical maximum value of $\mathcal{L}_{sec}$ is 4, and we set its weight to 1/4. However, normalizing the loss terms cannot lead to better performance. Thus, we keep the default setting which equally weigh each loss term by 1.0.
\begin{table}[ht]
    \centering
    \caption{Comparison between different loss weighting strategies.}
    \label{tab:loss_norm}
    \begin{tabular}{l|cc}
    \toprule
         Loss Weights & Dev & Test \\
    \midrule
         Normalization & \textbf{20.5} & 20.6 \\
         All 1.0 (default) & \textbf{20.5} & \textbf{20.4} \\
    \bottomrule
    \end{tabular}
\end{table}

\subsection{Ablation Study for Local Transformer}
We conduct an ablation study on Phoenix-2014 to verify the effectiveness of the Gaussian bias and the DTCN layer. As shown in Table \ref{tab:local_tf}, both the Gaussian bias and the DTCN layer can significantly decrease WER, proving that our local transformer is a strong sequential module for CSLR.
\begin{table}[ht]
    \centering
    \caption{Ablation study for the local transformer. (GB: Gaussian bias; DTCN: depth-wise temporal convolution network.)}
    \label{tab:local_tf}
    \begin{tabular}{cc|c}
    \toprule
         GB & DTCN & WER\%  \\
    \midrule
          &  & 25.2 \\
         \checkmark &  & 22.7 \\
         \checkmark & \checkmark & \textbf{21.5} \\
    \bottomrule
    \end{tabular}
\end{table}

\subsection{Position of the Signer Removal Module}
The position of the signer removal module (SRM) is simply a hyper-parameter. Besides SRM, the position of the spatial attention module is also empirically decided. As shown in Table \ref{tab:position_sam}, we first decide the position of the spatial attention module by putting it after different CNN layers. We find that an intermediate position, \ie, after the 5th CNN layer, is the best choice. An intuitive explanation is that early positions cannot provide enough error signals for the visual module, while a position too late will lead to low heatmap resolution. After that, we fix the position of the spatial attention module and adjust the position of the SRM. As shown in Table \ref{tab:position_srm}, putting the SRM right after the 5th CNN layer is also the best choice.

\begin{table}[ht]
\centering
\caption{The effects of the position of the spatial attention module and signer removal module. $m$ denotes the index of the CNN layer. All experiments are conducted on PHOENIX-2014-SI.}
\label{tab:position}
\begin{subtable}[t]{1.0\textwidth}
\centering
\caption{The effect of the position of the spatial attention module.}
\label{tab:position_sam}
\begin{tabular}{l|cccccccc}
    \toprule
    $m$ & 1 & 2 & 3 & 4 & 5 & 6 & 7 & 8 \\
    \midrule
    Resolution & 224 & 112 & 56 & 56 & 28 & 28 & 14 & 14 \\
    Dev & 36.1 & 34.9 & 35.5 & 36.1 & \textbf{34.3} & 35.5 & 36.2 & 37.2 \\
    Test & 36.0 & 35.3 & 35.5 & 36.5 & 34.4 & \textbf{33.6} & 35.7 & 36.0 \\
    \bottomrule
\end{tabular}
\end{subtable}

\begin{subtable}[t]{1.0\textwidth}
\centering
\caption{The effect of the position of the signer removal module.}
\label{tab:position_srm}
\begin{tabular}{l|cccccccc}
    \toprule
    $m$ & 1 & 2 & 3 & 4 & 5 & 6 & 7 & 8 \\
    \midrule
    Dev & 35.3 & 35.3 & 36.2 & 35.3 & \textbf{33.1} & 35.2 & 35.9 & 35.1 \\
    Test & 35.3 & 34.4 & 35.0 & 35.0 & \textbf{32.7} & 35.6 & 33.2 & 34.1 \\
    \bottomrule
\end{tabular}
\end{subtable}
\end{table}

\subsection{Abbreviations}
We list all abbreviations appeared in main texts and their corresponding full names in Table \ref{tab:abbr}.
\begin{table}[ht]
    \centering
    \caption{Abbreviations and full names.}
    \label{tab:abbr}
    \resizebox{\linewidth}{!}{
    \begin{tabular}{c|c|c|c}
    \toprule
         Abbreviation & Full Name & Abbreviation & Full Name \\
    \midrule
         BiLSTM & bidirectional long short-term memory & RNN & recurrent neural network \\
         CMP & channel-wise max pooling & SAC & spatial attention consistency \\
         CNN & convolutional neural network & SD & signer-dependent \\
         CSLR & continuous sign language recognition & SEC & sentence embedding consistency \\
         $\text{C}^2$SLR & consistency-enhanced CSLR & SEE & sentence embeddding extractor \\
         CT & CNN+TCN & SFD & stochastic frame dropping \\
         CTC & connectionist temporal classification & SI & signer-independent \\
         GAP & global average pooling & SLR & sign language recognition \\
         GB & Gaussian bias & SOTA & state-of-the-art \\
         GFE & gloss feature enhancement & SP & statistical pooling \\
         GR & gradient reversal & SRM & signer removal module \\
         IR & informative region & TCN & temporal convolutional network \\
         ISLR & isolated sign language recognition & VA & visual alignment \\
         LSA & local self-attention & VE & visual enhancement \\
         LT & local transformer & VLT & VGG11+LT \\
         MLP & multi-layer perceptron & VTB & VGG11+TCN+BiLSTM \\
         QK & query-key & WER & word error rate \\
         ReLU & rectified linear unit & & \\
    \bottomrule
    \end{tabular}
    }
\end{table}

\end{document}